\pdfoutput=1
\documentclass[lettersize,journal]{IEEEtran}
\usepackage{amsmath,amsfonts}
\usepackage{array}
\usepackage{stfloats}
\usepackage{url}
\usepackage{cite}
\usepackage{graphicx}
\usepackage{textcomp}
\usepackage{xcolor}
\usepackage[english]{babel}
\usepackage{acronym}
\usepackage{mathtools}
\usepackage{booktabs}
\usepackage{changes}
\DeclareMathOperator*{\argmax}{arg\,max}
\usepackage{caption}
\usepackage{subcaption}
\usepackage[ruled]{algorithm2e}
\usepackage{pgfplots}
\usepackage{pifont}% http://ctan.org/pkg/pifont
\usepackage{multirow}
\usepackage{makecell}
\newcommand{\cmark}{\ding{51}}%
\newcommand{\xmark}{\ding{55}}%

\begin{document}

\title{Autonomous Search of Semantic Objects in Unknown Environments}

\author{Zhentian Qian$^{1}$, Jie Fu$^2$, and Jing Xiao$^{1}$% <-this % stops a space
%\thanks{*This work was not supported by any organization}% <-this % stops a space
\thanks{$^{1}$Zhentian Qian and Jing Xiao are with the Robotics Engineering Department, Worcester Polytechnic Institute, Worcester, MA, USA 
        {\tt\small zqian@wpi.edu, jxiao2@wpi.edu }}%
\thanks{$^{2}$Jie Fu is with the Department of Electrical and Computer Engineering, University of Florida, Gainesville, FL, USA
        {\tt\small fujie@ufl.edu}}%
}

\markboth{Journal of \LaTeX\ Class Files,~Vol.~14, No.~8, August~2021}%
{Shell \MakeLowercase{\textit{et al.}}: Autonomous Search of Semantic Objects in Unknown Environments}

\IEEEpubid{0000--0000/00\$00.00~\copyright~2021 IEEE}

\maketitle

\begin{abstract}
This paper addresses the problem of enabling a robot to search for a semantic object, i.e., an object with a semantic label, in an unknown and GPS-denied environment. For the robot in the unknown environment to detect and find the target semantic object, it must perform simultaneous localization and mapping (SLAM) at both geometric and semantic levels using its onboard sensors while planning and executing its motion based on the ever-updated SLAM results. In other words, the robot must be able to conduct simultaneous localization, semantic mapping, motion planning, and execution in real-time in the presence of sensing and motion uncertainty. 
%in partially observable environments. 
This is an open problem as it combines semantic SLAM based on perception and real-time motion planning and execution under uncertainty. Moreover, the goals of the robot motion change on the fly depending on whether and how the robot can detect the target object. We propose a novel approach to tackle the problem, leveraging semantic SLAM,  Bayesian Networks, Markov Decision Process, and Real-Time Dynamic Programming. The results in simulation and real experiments demonstrate the effectiveness and efficiency of our approach.
\end{abstract}


\section{Introduction}
% Talk about the need in the real world to search semantic objects in an unknown environment 
\label{sec: intro}
\IEEEPARstart{T}his paper is motivated by the problem of searching an unknown and GPS-denied environment for some target object by a robot in a timely fashion, which is a fundamental problem in many robotics application scenarios, from search and rescue to reconnaissance to elderly care. For example, rescue robots are deployed at an earthquake site to search for survivors where the environment is complicated and unknown, and time is a matter of life and death.

% and the challenges to do so.
Despite the importance of this problem, it remains largely unsolved as there are three major challenges in this problem:
\begin{itemize}
    \item The environment is unknown and GPS-denied (such as inside an unknown building) so that no map, neither in a geometrical nor in a semantic sense, is provided to the robot. 
    \item The target object's location is unknown to the robot. 
    \item The robot has to rely on limited and noisy sensing input to perceive the environment as it moves in the presence of motion uncertainty, resulting in estimation errors in mapping and robot localization. 
\end{itemize}

%Explain what characteristics are needed for a robotic system to solve the problem of searching semantic objects. Those characteristics will appear later as columns of (to be revised) Table 1. 
To address those challenges, the robot needs to have the following capabilities:

\smallskip
\noindent 
{\bf SLAM}: the robot must perceive the environment to build geometric and semantic maps and localize itself.

\smallskip
\noindent 
{\bf Exploration}: since the environment is unknown, the robot must have autonomous exploration capabilities.

\smallskip
\noindent 
{\bf Semantic prior knowledge usage}: the robot should be able to use semantic prior knowledge related to the target object to make the search opportunistic and efficient.

\smallskip
\noindent 
{\bf Probabilistic planning}: the robot must plan its motion in the presence of both perception and motion uncertainty to maximize the probability of completing the task. 

\smallskip
\noindent 
{\bf Adaptation}: as the robot explores the environment, it should not only expand but improve map building (by reducing perception uncertainty), and with the expanded and improved map, its search motion plan for the target object should be improved also. 

\smallskip
\noindent 
{\bf Uncertainty reduction}: the robot must be able to improve semantic map building by reducing map uncertainty through its action, which in turn enables the robot to find the target object more efficiently. 

\smallskip
\noindent 
{\bf Mission status awareness}: the robot should be able to perceive its current mission progress. That is, the robot should know whether and with what confidence score it has completed the task. 

\smallskip
\noindent 
{\bf Goal creation on the fly}: the robot must change its goal of motion on the fly, depending on its perception of the environment to decide if it should explore, determine and move towards an intermediate goal relevant to the target object, or move to reduce perception uncertainty of the semantic map (including the target object).

%T Point out that related existing work is compartmentalized and focuses on certain aspects of the overall problem while making assumptions about the other aspects. 
Related existing methods are compartmentalized to focus only on some of the above capabilities while making assumptions about the rest, as discussed in more depth in section \ref{sec: related work}. How to enable a robot to possess all the key capabilities is necessary for solving the problem but is not addressed in the current literature. 
\IEEEpubidadjcol
\section{Related Work}
\label{sec: related work}
We will review related work in simultaneous localization and mapping, and robot path and motion planning, including next-best view planning below. 
\subsection{SLAM}
There is a significant amount of literature on simultaneous localization and mapping (SLAM) \cite{thrun1998probabilistic, DurrantWhyte1996LocalizationOA} for robot mapping and navigation in an unknown environment based on perception, such as visual and odometry sensing \cite{fraundorfer2011visual}. SLAM methods model and reduce sensing uncertainties in mapping an unknown environment and localizing the robot in it at the same time. Semantic SLAM and active SLAM are particularly relevant. 
%As such, this problem has received attention from both the SLAM community and the planning community, each with its own emphasis and limitations. 
%The problem is mainly studied in the scope of active SLAM and semantic SLAM. 
\subsubsection{Semantic SLAM}
Semantic SLAM methods are focused on representing, mapping, and localizing 3D objects using different representations of objects such as meshes \cite{galvez2016real}, quadric \cite{nicholson2018quadricslam, qian2020semantic, 9691853}, cuboid \cite{yang2019cubeslam}, and OctoMap \cite{zhang2018semantic}.
\subsubsection{Active SLAM}
Active SLAM \cite{feder1999adaptive} aims to choose the optimal trajectory for a robot to improve map and localization accuracy and maximize the information gain. The localization accuracy is typically measured by metrics such as A-opt (sum of the covariance matrix eigenvalues)\cite{leung2006active, kollar2008trajectory}, D-opt (product of covariance matrix eigenvalues) \cite{kim2013perception}, E-opt (largest covariance matrix eigenvalue) \cite{ehrenfeld1955efficiency}. Information gain is measured in metrics such as joint entropy \cite{stachniss2005information} and expected map information \cite{blanco2008novel}. 

However, neither semantic nor active SLAM 
%considers the navigation tasks and is certainly not optimized for the abovementioned problem. 
considers performing tasks other than mapping an unknown environment. The planning aspect is not addressed for semantic SLAM and is downplayed in active SLAM with simple methods such as A*\cite{kim2013perception}. 

\subsection{Planning}
Robot path and motion planning is one of the most studied areas in robotics. The basic objective is to find an optimal and collision-free path for a robot to navigate to some goals in an environment. Many traditional path-planning approaches assume a more or less known environment, i.e., the robot already has a map and models of objects \cite{lavalle2006planning}. On the other hand, real-time, sensing-based planning in an unknown environment remains largely a challenge \cite{alterovitz2016robot}. 

Earlier work includes grid-based planning approaches such as D* \cite{stentz1997optimal} and D* Lite \cite{koenig2005fast}, sampling-based approaches such as ERRT\cite{bruce2002real} and DRRT \cite{ferguson2006replanning}, and adaptive approaches such as \cite{RAMP}. These approaches consider the environment dynamic and only partially known, but assume the goal position is known, disregard the uncertainties in sensing, the robot pose, and dynamics, and do not consider semantic information. 

Recently, various techniques based on partially observable Markov decision processes (POMDPs) have been developed \cite{wang2022hybrid, veiga2019hierarchical, burks2019optimal} to incorporate sensing and robot motion uncertainties into planning in partially observable environments. However, POMDP suffers from the curse of dimensionality and is computationally expensive, particularly when the state space is large. For the POMDP to scale, high-level abstraction must be made for the state space. For example, treat objects \cite{veiga2019hierarchical} or rooms\cite{wang2022hybrid} as state variables. The downside is that highly abstracted models can lose touch with reality. To bypass this problem, some researchers turn to deep learning to learn semantic priors and make predictions on the unobserved region (PONI\cite{ramakrishnan2022poni}, SemExp\cite{chaplot2020object}, L2M\cite{georgakis2021learning}). These methods tend to suffer from poor generalization. 

Next-best view planning \cite{connolly1985determination} is another highly related topic, designed for efficient visual exploration of unknown space. Unlike active SLAM, approaches for next-best view planning typically do not consider robot localization uncertainty. A next-best view planner starts by sampling a set of views in the environment, evaluates the estimated information gain for each view, and selects the view with the maximum information gain as the next view \cite{zeng2020view}. Different approaches differ in the sampling methods (uniform sampler, frontier-based coverage sampler \cite{meng2017intelligent}), information gain (path costs are incorporated in \cite{selin2019efficient, meng2017intelligent}), and the selection of the next view (such as receding horizon scheme in \cite{bircher2016receding} and Fixed Start Open Traveling Salesman Problem (FSOTSP) solver in \cite{meng2017intelligent}).

However, existing planning methods in unknown environments usually do not consider real-time results from SLAM with embedded and changing uncertainties, such as the robot's pose, the metric map, and the semantic map (generated by semantic SLAM). Only the metric map was used by next-best view planning approaches \cite{zeng2020view, meng2017intelligent, selin2019efficient}. 
%The ever-changing nature of SLAM results is also not properly treated. 
%Even the next-best view planning which is based on the updated SLAM mapping results, only the metric  map is utilized in this process. 

% path and motion planning methods usually do not consider the SLAM problem associated with mapping and localization in unknown environments. Even for those that do \cite{kantaros2022perception}, SLAM is only treated as an abstract model rather than a system running in parallel.

% what's the gap?
% On the other hand, \cite{kantaros2022perception} mitigated this issue by assuming deterministic dynamics, using atomic predicates to hard-threshold the uncertainty of the map, and applying a sampling-based algorithm.

\subsection{Limitations}
We further discuss the limitations of the existing work when applied to the problem studied in this paper, summarized in Table \ref{tab:comp}. Specifically, we evaluate each method for the essential capabilities identified in Section \ref{sec: intro}. For example, SLAM approaches only have the SLAM capability. Active SLAM and next-best view planning methods do not employ semantic SLAM. They have exploration capability and limited adaption capability as plans can be changed based on the current map. These methods take explicit actions to reduce the map and pose uncertainty by creating and reaching a set of intermediate goals. 

Traditional grid-based planning approaches such as D*, D* Lite, and sampling-based planning approaches such as ERRT and DRRT can also adapt to changes of objects in the map, but do not consider uncertainties. POMDP-based approaches are intended for probabilistic planning by taking into account uncertainties and maintaining belief about the mission status. However, these approaches do not incorporate SLAM, are compartmentalized, and miss essential capabilities (Section \ref{sec: intro}) required for our overarching problem. 

Several deep-learning based approaches exhibit more comprehensive capabilities. Nevertheless, PONI and SemExp only have mapping but not localization capabilities as they rely on ground-truth robot pose information. The uncertainty in the map, robot motion, and robot pose estimation are not treated when formulating plans to reach intermediate goals. No explicit action is taken to reduce map uncertainty. Although L2M improves upon those methods by taking uncertainty reduction into goal formulation, it still misses two capabilities either partially or entirely. 

\begin{table*}[htbp]
    \caption{Related existing work vs. our method in this paper}
    \centering
    \begin{tabular}{c c c c c c c c c}
    \toprule
        Method & SLAM & Exploration & \makecell{Semantic\\ prior} & \makecell{Probabilistic\\ planning} & Adaptation & \makecell{Uncertainty\\ reduction} & \makecell{Mission status\\ awareness} & \makecell{Goal \\ creation} \\
    \midrule 
 SLAM &  \cmark & \xmark & \xmark & \xmark & \xmark & \xmark & \xmark & \xmark\\
    Active SLAM & No semantics & \cmark & \xmark & \xmark & \cmark & \cmark & \xmark & \cmark \\
Next-best view & No semantics & \cmark & \xmark & \xmark & \cmark & \cmark & \xmark & \cmark\\
    D*, D* Lite &  \xmark & \xmark & \xmark & \xmark & \cmark & \xmark & \xmark & \xmark\\
ERRT, DRRT &  \xmark & \xmark & \xmark & \xmark & \cmark & \xmark & \xmark & \xmark\\
POMDP & \xmark & \xmark & \xmark & \cmark &  \xmark & \xmark & \cmark & \xmark\\
    PONI   &  No localization & \cmark & \cmark & \xmark & \cmark & \xmark & \cmark & \cmark\\
    SemExp    & No localization & \cmark & \cmark & \xmark & \cmark & \xmark & \cmark & \cmark\\
       L2M    & No localization & \cmark & \cmark & \xmark & \cmark & \cmark & \cmark & \cmark\\
       \textbf{Our method} &  \textbf{\cmark} & \textbf{\cmark} & \textbf{\cmark} & \textbf{\cmark} & \textbf{\cmark} & \textbf{\cmark} & \textbf{\cmark} & \textbf{\cmark} \\
    \bottomrule
    \end{tabular}
    \label{tab:comp}
\end{table*}

\section{Approach and Contributions}
\label{sec: task}
% Section 3: Approach and contribution 
% This section summarizes our approach and contributions to indicate that our approach has all the desired characteristics (i.e., indicated in Table I) and fill the gaps that the existing approaches have. 

% It further introduces the rest of the sections in this paper. 

%Aware of the strength and limitations of the work conducted in the SLAM and planning communities,  
\begin{figure}
    \centering
    \includegraphics[width=\linewidth]{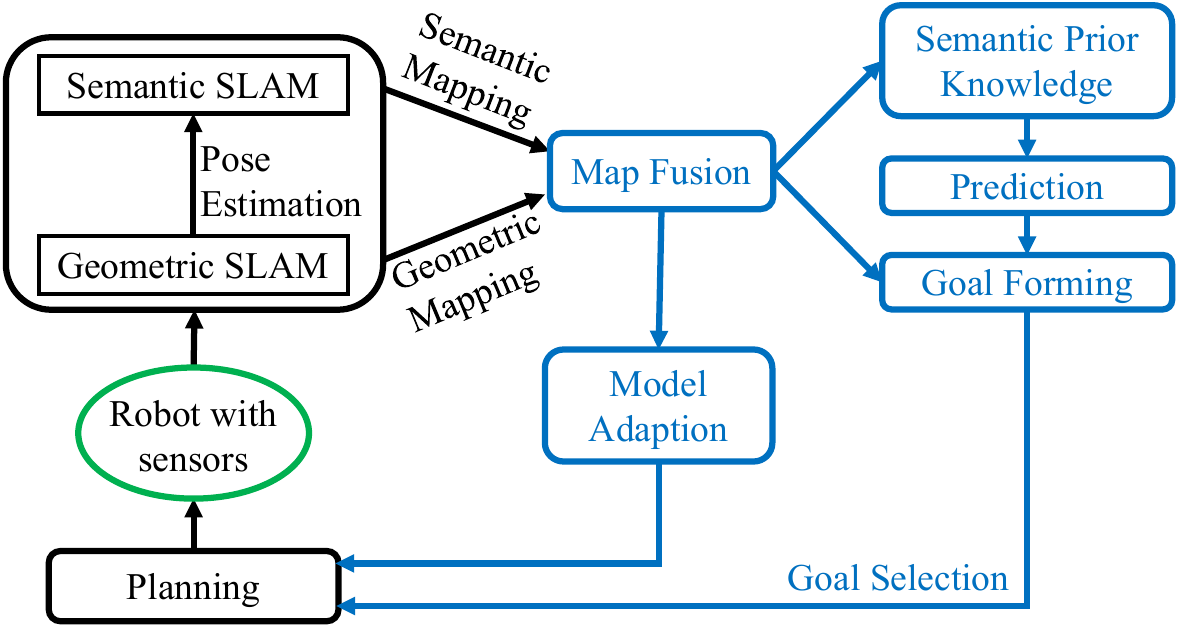}
    \caption{Overview of the online system. Novel parts are in blue.}
    \label{fig:sys}
\end{figure}

This paper introduces a novel, adaptive system that equips the robot with all the eight essential capabilities listed in Section \ref{sec: intro}. 

An overview of our online, real-time approach is shown in  Fig. \ref{fig:sys}, where the novel contributions of this paper are highlighted in blue. The geometric SLAM and the custom semantic SLAM modules generate and update the robot pose estimation, a  geometric map, and a semantic map. Using semantic prior knowledge, our system predicts the target object's position based on the current fused map. The predicted position, together with the fused map, are fed into the goal-forming module. This module selects an intermediate goal and passes it into the planning module. The planning module uses a probabilistic planner to account for all uncertainties.  
%and adapts to the ever-updated SLAM results. 
Finally, as the generated plan is being executed by the robot, the SLAM modules update the map and pose estimation based on the expanded sensing information, which in turn is used to update the fused map and adapt the models of the environment and robot for the planner, forming a closed-loop system (with two loops). 

Compared to existing approaches, our system's main contributions include: 
\begin{enumerate}
    \item incorporating geometric SLAM and custom semantic SLAM modules to enable full SLAM capabilities for the robot;
    \item incorporating prior semantic knowledge to facilitate the search of semantic objects; 
    \item on-the-fly goal selection enabling the robot to perform exploration, more opportunistic search based on semantic objects detected, 
    %the predicted object location, 
    to reduce map uncertainty, or to complete the task when the target object is considered found and the mission is accomplished;
    \item probabilistic planning that accounts for the uncertainties in pose estimation, map, and robot motion and adapts the models  for producing the policy for the next robot motion based on the ever-updated SLAM results; 
    \item interconnections between modules so that each module is mutually dependent and mutually facilitating the related modules.
\end{enumerate}
Finally, the fused maps produced in the autonomous search of a target object with our system can be used for future tasks in the same environment, and each time, the maps will be enhanced in terms of accuracy and coverage (as the robot explores more of the environment with semantic SLAM) to enable more efficient and effective robot performance in accomplishing tasks. 

\section{Mapping and localization}
In this section, we describe how geometric and semantic SLAM is achieved, and the information from different levels is fused into a single map $E_t$. 

\subsection{Geometric SLAM}
\begin{figure}
    \centering
    \includegraphics[width =0.6\linewidth]{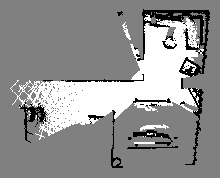}
    \caption{An example grid map at time $t$. The grey, white, and black areas represent unknown, free, and occupied regions.}
    \label{fig:grid map}
\end{figure}

Geometric SLAM is run in parallel with semantic SLAM in our system. 
We employ the RTAB-MAP \cite{labbe2019rtab} algorithm for geometric SLAM. It generates a grid map $\mathbf{G}_t \in \{0, 1, -1\}^{W\times H}$, where $W$ and $H$ are the width and height of the grid map. $0$, $1$, and $-1$ in the grid map represent free, occupied, and unknown space respectively, as shown in Fig. \ref{fig:grid map}. The geometric SLAM module also estimates the robot pose $(\mathbf{\mu}_{p,t}, \mathbf{\Sigma}_{p,t})$, where $\mathbf{\mu}_{p, t}$ and $\mathbf{\Sigma}_{p, t}$ are the mean and covariance of the robot pose at time instance $t$.

\subsection{Semantic SLAM}
We adapt the system introduced in our previous work \cite{qian2020semantic} for semantic SLAM. At time instance $t-1$, the position estimation $\mathbf{m}_{i, t-1} \in \mathbb{R}^2$ for the semantic object $o_i$  is:
\begin{equation}
    bel(\mathbf{m}_{t-1}) \sim \mathcal{N}(\mu_{t-1}, \mathbf{\Sigma}_{t-1}),
    \label{eq: map pos dis}
\end{equation}
Where $bel(\cdot)$ stands for the belief over a variable. Note that for simplicity, the subscript $i$ is dropped in \eqref{eq: map pos dis}, as in \eqref{eq: range bearing}--\eqref{eq: Bayes theorem object class}.
%Our analysis would revolve around a single object $o_i$ and can be easily extended to multiple objects.  

At time instance $t$, the robot pose $\mathbf{x}_{t} \in \mathbb{R}^2$ estimated by geometric SLAM is $bel(\mathbf{x}_{t}) \sim \mathcal{N}(\mathbf{\mu}_{p,t}, \mathbf{\Sigma}_{p,t})$.

If the semantic object $o_i$ is detected on the color image $\mathbf{I}_t$, range-bearing measurement $\mathbf{z}_t$ will be generated based on the depth information of $o_i$ from the depth image. The range-bearing measurement noise $\mathbf{\delta}_t$ is:  
\begin{equation}
    \mathbf{\delta}_t \sim \mathcal{N}(0, \mathbf{\Sigma}_{\delta}).
    \label{eq: range bearing}
\end{equation}
The covariance of the range-bearing measurement $\Sigma_\delta$ is assumed to be time-dependent. Then the posterior belief $bel(\mathbf{m}_t)$ at time $t$ can be updated using Bayes' theorem:
\begin{equation}
\begin{split}
      bel(\mathbf{m}_t) & = p(\mathbf{m} \vert \mathbf{z}_{1:t}) = \frac{p(\mathbf{z}_{t} \vert \mathbf{m}, \mathbf{z}_{1:t-1}) \cdot p(\mathbf{m} \vert \mathbf{z}_{1:t-1})}{p(\mathbf{z}_t \vert \mathbf{z}_{1:t-1})} \\
      & = \eta \int p(\mathbf{z}_{t} \vert \mathbf{m}, \mathbf{x}_t)\cdot bel(\mathbf{x}_t)\cdot bel(\mathbf{m}_{t-1}) d\mathbf{x}_t,
\end{split}
\label{eq:filter}
\end{equation}
where $\eta$ is a normalizing term. 

Substituting the probability density functions of $p(\mathbf{z}_{t} \vert \mathbf{m}, \mathbf{x}_t)$, $ bel(\mathbf{x}_t)$, and $bel(\mathbf{m}_{t-1})$ into \eqref{eq:filter}, the final result after simplification suggests that the updated posterior belief $bel(\mathbf{m}_{t})$ can be approximated by a multivariate Gaussian distribution $bel(\mathbf{m}_{t}) \sim \mathcal{N}(\mu_{t}, \mathbf{\Sigma}_{t})$: 
\begin{align*}
    \mathbf{\Sigma}_{t} & =  \Big(\mathbf{K}_1^T\mathbf{\Sigma} _{\delta}^{-1}\mathbf{K}_1 + \mathbf{\Sigma} _{t-1}^{-1} - \mathbf{K}_1^T\mathbf{\Sigma} _{\delta}^{-1}\mathbf{K}_2 \mathbf{\Psi}  \mathbf{K}_2^T\mathbf{\Sigma} _{\delta}^{-1}\mathbf{K}_1\Big)^{-1},\\
     \mu_{t} & = \mu_{t-1} +  \mathbf{\Sigma}_{t} \mathbf{K}_1^T(\mathbf{\Sigma} _\delta^{-1} - \mathbf{\Sigma} _\delta^{-1}\mathbf{K}_2\mathbf{\Psi}\mathbf{K}_2^T\mathbf{\Sigma} _\delta^{-1})\Delta\mathbf{z}_t.
\end{align*}
where $\Delta\mathbf{z}_t$ is the error between expected and actual range-bearing measurement,
\begin{equation*}
\mathbf{\Psi}^{-1}= \mathbf{K}_2^T\mathbf{\Sigma} _\delta^{-1}\mathbf{K}_2 + \mathbf{\Sigma} _p^{-1},
\end{equation*}
\begin{equation*}
\mu_{p, t} = \begin{bmatrix} \mu_x, \mu_y, \mu_\theta\end{bmatrix}^T,
\end{equation*}
\begin{equation*}
\mu_{t} = \begin{bmatrix} \mu_{m,x}, \mu_{m,y}\end{bmatrix}^T,
\end{equation*}
\begin{equation*}
\mathbf{K}_1 = \begin{bmatrix}
\frac{\mu_{m,x} - \mu_x}{\sqrt{(\mu_{m,x} - \mu_x)^2 + (\mu_{m,y} - \mu_y)^2}} & \frac{\mu_{m,y} - \mu_y}{\sqrt{(\mu_{m,x} - \mu_x)^2 + (\mu_{m,y} - \mu_y)^2}} \\
\frac{\mu_{y} - \mu_{m, y}}{(\mu_{m,x} - \mu_x)^2 + (\mu_{m,y} - \mu_y)^2} & \frac{\mu_{m, x}- \mu_{x}}{(\mu_{m,x} - \mu_x)^2 + (\mu_{m,y} - \mu_y)^2} 
\end{bmatrix},
\end{equation*}

\begin{equation*}
\mathbf{K}_2 = \begin{bmatrix}
\frac{\mu_x - \mu_{m,x} }{\sqrt{(\mu_{m,x} - \mu_x)^2 + (\mu_{m,y} - \mu_y)^2}} & \frac{\mu_{m, y} - \mu_{y}}{(\mu_{m,x} - \mu_x)^2 + (\mu_{m,y} - \mu_y)^2} \\
\frac{\mu_y- \mu_{m,y} }{\sqrt{(\mu_{m,x} - \mu_x)^2 + (\mu_{m,y} - \mu_y)^2}} & \frac{\mu_{x}-\mu_{m, x}}{(\mu_{m,x} - \mu_x)^2 + (\mu_{m,y} - \mu_y)^2} \\
0 & -1 \\
\end{bmatrix}^T.
\end{equation*}

The object class probability distribution $p_{t}(\cdot)$ is updated using Bayes' rule:
\begin{equation}
\begin{split}
p_{t}(c) & = p(c \vert \mathbf{L}_{1:t}) =  \frac{p(\mathbf{L}_{t} \vert c, \mathbf{L}_{1:t-1}) \cdot p(c \vert \mathbf{L}_{1:t-1})}{p(\mathbf{L}_t \vert \mathbf{L}_{1:t-1})} \\
&=  \eta p(\mathbf{L_t}\lvert c) \cdot p_{t-1}(c) = \frac{p(\mathbf{L_t}\lvert c) \cdot p_{t-1}(c)}{\sum_{c' \in \mathbb{C}} p(\mathbf{L}_t\lvert c') p_{t-1}(c')},    
\end{split}
\label{eq: Bayes theorem object class}
\end{equation}
where $\eta = 1/p(\mathbf{L}_t \vert \mathbf{L}_{1:t-1})$ is a normalization constant, $\mathbf{L}_t \in \mathbb{R}^{\lvert \mathbb{C} \rvert}$ is the confidence level distribution of an object in different classes, returned by an object detector, such as YOLOv3 \cite{redmon2018yolov3} at time $t$. $c \in \mathbb{C}$ is one of the possible object classes. 
$p(\mathbf{L}_t\vert c)$ is the object detector uncertainty model, representing the probability of object detector outputs $\mathbf{L}_t$ when the object class is $c$. We use the Dirichlet distribution ${\displaystyle \operatorname {Dir} ({\boldsymbol {\alpha }}_c)}$ to model this uncertainty, with a different parameter $\boldsymbol{\alpha}_c \in \mathbb{R}^{\lvert \mathbb{C} \rvert}$ for each object class $c$. 

\subsection{Map Fusion}
\begin{figure}
    \centering
    \includegraphics[width =0.6\linewidth]{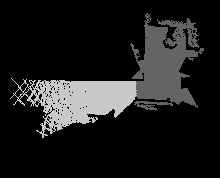}
    \caption{Segmented geometric room examples at time $t$. The two segmented rooms are encoded in different colors.}
    \label{fig:segmented room}
\end{figure}

As the robot explores an environment, our system uses off-the-shelve tools \cite{bormann2018indoor} to segment the gradually built grid map into different geometric rooms: a room is defined as any space enclosed within a number of walls to which entry is possible only by a door or other dividing structure that connects it either to a hallway or to another room. Every grid on the grid map $\mathbf{G}_t$ is assigned with a corresponding room ID: $\mathbf{R}_t \in \mathbb{N}^{W\times H}$. An example is provided in Fig. \ref{fig:segmented room}.

Based on the locations of map objects, the corresponding geometric room IDs are assigned to the objects. Formally, a map object $o_{i}$ is represented as a 4-tuple $o_{i} = \langle \mu_{i}, \Sigma_{i}, p_{i}, r_{i} \rangle$ with $\mu_i$ and $\Sigma_i$ the mean and covariance of the object $o_i$ pose, $p_i$ the object class distribution, and $r_{i}$ the room ID of $o_{i}$. The object map is the set of observed map objects $\mathbb{O}_t = \{o_1, o_2, \ldots, o_n\}$. 

Finally, the fused map $E_t = \langle\mathbf{G}_t, \mathbb{O}_t, \mathbf{R}_t \rangle$ collects the grid map $\mathbf{G}_t$, object map $\mathbb{O}_t$, as well as the room information $\mathbf{R}_t$. 

\section{Semantic Prior Knowledge}
\label{sec: semantic prior}
Our robot planner leverages prior semantic knowledge to facilitate efficient exploration. The key idea is that objects in the target category may have a closer affinity to some categories of objects than others. The co-occurrence relationship between objects of two categories is estimated based on Lidstone's law of succession \cite{schutze2008introduction}:
\begin{equation}
    p(c_i \mid c_j) = \frac{N(c_i, c_j) + \alpha}{N(c_j) + \alpha \lvert \mathbb{C} \rvert},
    \label{eq: lidstone}
\end{equation}
where $p(c_i \mid c_j)$ is the conditional probability of objects of class $c_i$ being in a geometric room given objects of class $c_j$ is already observed in the same room. $N(c_i, c_j)$ is the number of times objects of classes $c_i$ and $c_j$ are observed in the same room. $N(c_j)$ is the number of times objects of category $c_j$ are observed in a room. $\alpha \in [0, \infty)$ is a smoothing factor, and finally $\lvert \mathbb{C} \rvert$ is the number of classes. 
 
\begin{figure}
    \centering
    \includegraphics[width=1\linewidth]{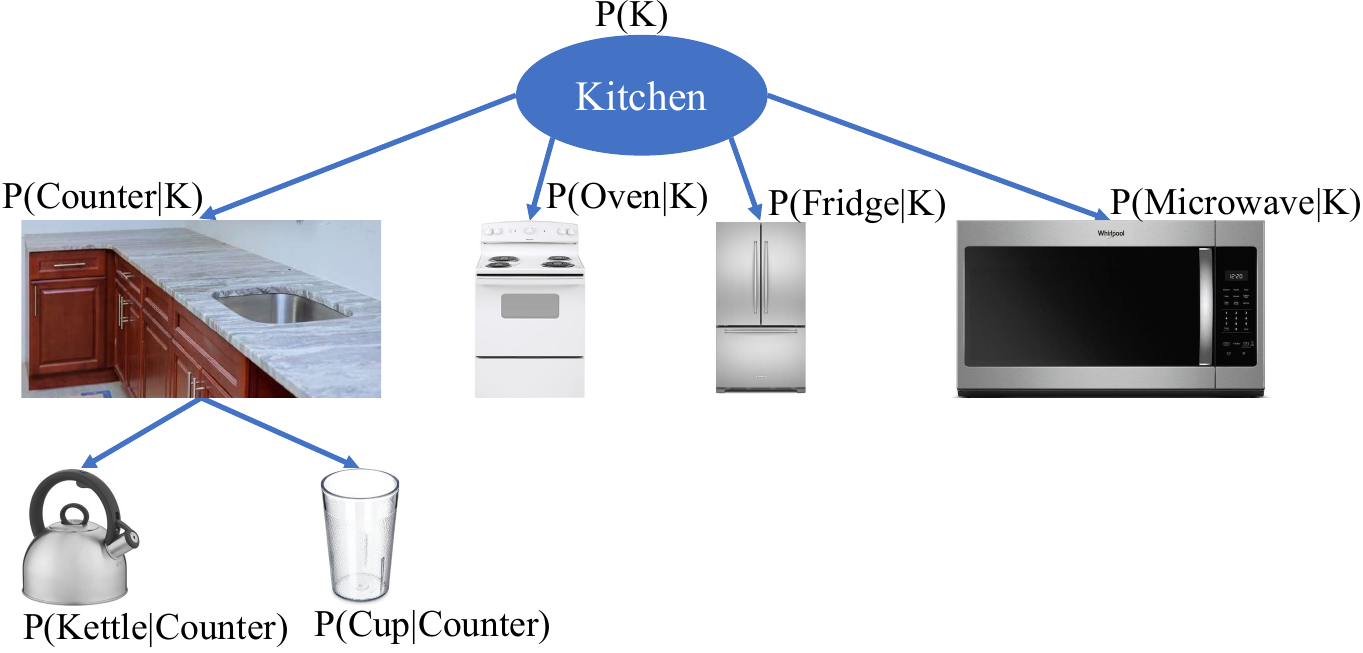}
    \caption{Example Bayesian network for a kitchen environment.}
    \label{fig:bayesian net}
\end{figure}

The probabilistic co-occurrence relationships of multiple pairs of objects are captured using Eq. \eqref{eq: lidstone} and further assembled into multiple Bayesian networks. We construct a set of Bayesian networks $\mathcal{B} = \{B_1, B_2, \ldots \}$, with one for each semantic space $\mathcal{S} = \{S_1, S_2, \ldots \}$. Each semantic space corresponds to one room category, such as kitchen, office, bathroom, etc. An example of a Bayesian Network is illustrated in Fig. \ref{fig:bayesian net}, demonstrating common object classes found in a kitchen,  and their conditional dependency. 

For each geometric room $r_i$ in the environment, we will collect the set of object classes $\mathbb{E}_i = \{c_1, c_2, \ldots \}$ that are observed in the room $r_i$. Recall that we keep a class probability distribution for each map object. Thus we cannot draw a deterministic conclusion regarding the presence of a certain object class in the room $r_i$. However, to keep the problem tractable, we assume the presence of object class $c_k$ if for any object $o_j$ in the room, the probability of the object $o_j$ being in class $c_k$ exceeds some threshold $\lambda$: $c_k \in \mathbb{E}_i \iff \exists j,\, p_j(c_k) > \lambda$.

Given the evidence set $\mathbb{E}_i$, we only consider the Bayesian networks in $\mathcal{B}$ that contains the target object instance $c_T$ and contains some objects in $\mathbb{E}_i$; name this subset of Bayesian networks $\mathcal{B}_i$. By doing so, we narrow down the possible semantic space categories for the geometric room $r_i$ to a subset $\mathcal{S}_i$, which corresponds to $\mathcal{B}_i$. For each Bayesian network $B_j \in \mathcal{B}_i$, our system can compute the probability of finding the target object instance $o_T$ in the room $r_i$ based on evidence $\mathbb{E}_i$, denoted as $P(c_T \mid \mathbb{E}_i , r_i, B_j)$.

Our robot system can then infer the probability of finding the target object instance $o_T$ in the same room $r_i$ by feeding $\mathbb{E}_i$ into the Bayesian network set $\mathcal{B}_i$:
\begin{equation*}
     P(c_T \mid \mathbb{E}_i = \{c_1, c_2, \ldots\}, r_i) = \max_{B_j \in \mathcal{B}_i}  P(c_T \mid \mathbb{E}_i , r_i, B_j).
\end{equation*}
This probability is computed for all geometric rooms.

%We would like to point out that should we choose to treat evidence as probabilistic, then we have to consider all Bayesian networks, and inference would be made on the power set of the evidence instead of the evidence itself:
%\begin{equation}
%     P(c_T \mid r_i) = \max_{B_j \in \mathcal{B}} \sum_{E\in 2^{\mathbb{E}_i}}  P(c_T \mid  E , r_i, B_j) P(E).
%\end{equation}
%The problem would soon become intractable. 

% Further, since room $r_i$ will be explored by means of frontier exploration, we can specify the probability of finding the target object instance $o_T$ by exploring a frontier edge $e_j$ inside room $r_i$:
% \begin{equation}
%     P(c_T \mid \mathbb{E}, e_j) \sim P(c_T \mid \mathbb{E}, r_i) \cdot \lvert e_j \rvert.
% \end{equation}
% The probability $P(c_T \mid \mathbb{E}, e_j)$ is designed also to be proportional to the size of the frontier edge $|e_j|$, factoring in the information gain by exploring $e_j$.

\section{Robot Modeling and Goal Forming}
In this section, we describe how an intermediate goal is determined on the fly for the robot and how the intermediate goal is incorporated into the robot model as the reward signal for subsequent planning.
\subsection{Robot Modeling}
To account for the uncertainty in robot pose estimation, SLAM map, and robot motion, the robot is modeled as a Markov decision process (MDP) $\text{M}_t = \langle  S, A,  s_0, P, R, F\rangle$ with the following components:\\
\noindent {---} $S$ is the discrete state space, representing the mean of the Gaussian distribution of the robot position. The mean of the robot's position is discretized and clipped to the closest grid cell in the grid map $\mathbf{G}_t$ to avoid an infinite state space.
	
\noindent {---} $A$ is a set of actions. We consider eight actions that allow the robot to move horizontally, vertically, and diagonally to reach its eight neighboring grid cells. A low-level controller maps the actions into the continuous robot actuation command.

\noindent {---} $s_0 \in S$ is the initial state. $s_0$ is determined by clipping the current robot position estimation $\mathbf{\mu}_{p,t}$ to the nearest grid
cell in the grid map $\mathbf{G}_t$.  
	
\noindent {---} $P\colon S \times A \times S \to [0, 1]$ is the transition probability function, where $P(\cdot \mid s, a)$ represents the probability distribution over next states given an action $a$ taken at the current state $s$. 
 For example, for the move-up action, the robot has a high probability of moving up one cell, but it also has a small probability of moving to the upper-left or upper-right cell due to uncertainty.  
	
\noindent {---} $R \colon S \times A \times S \to \mathbb{R}$ is the reward function, 
 %that maps the current prstate, action, and future state into a real value, 
 where $R(s, a, s')$ is the reward for executing action $a \in A$ at state $s \in S$ and reaching the next state $s' \in S$, which depends on the environment (captured by the fused map) and its changes after the robot's action $a$.
	
\noindent {---} $F \subset S$ is the set of (intermediate) target states, which are determined on the fly depending on the intermediate goal at the time, as described in Section \ref{sec: goal forming}. 

 %For the exploration task, if the robot corresponds to state $s$ is in obstacle or frontier region, $s \in F$. For the re-observation task, if the grid cell corresponds to state $s$ is in obstacle or visibility region $\mathbb{V}_t$, $s \in F$.

 \subsection{Goal Forming and Reward Design}
 \label{sec: goal forming}

As the robot's mission is to find a target object in an unknown environment, its goal of motion will be determined on the fly depending on the information provided by the fused map $E_t$ and the prediction made using semantic prior knowledge.  The mission is accomplished if the target object is visible and identified as such. 

There are several types of intermediate goals for the robot motion: 
\begin{enumerate}
\item If the target object is visible and identified as such, then the mission is accomplished and a stop signal is sent. 
\item If the target object is not included in the current map $E_t$, the robot chooses to explore where the target object likely is. This intermediate goal requires frontier detection. 
\item If an object in the map may be the target object (with a low probability
%threshold at 0.99 in experiments
), the robot chooses to observe more of the object in its visibility region and reduce the object's  uncertainty.
\end{enumerate}
At the beginning of each planning cycle, the robot evaluates the mission status and decides its intermediate goal. 
Next, the corresponding reward function of the robot MDP model $M_t$ is designed to drive the robot to a selected intermediate target state based on the intermediate goal. In the following, each type of intermediate goal and the corresponding reward function is described.  

\subsubsection{Mission Status for Task Completion}

To evaluate mission status, our robot system first identifies the object most likely in the target object category $c_T$. We refer to this object as the object of interest $o_{I}$, $I = \argmax_i p_i(c_T)$. If the probability of the object of interest $o_I$ being in the target object category $c_T$ exceeds a preset threshold $1 - \epsilon$, i.e., $p_I(c_T) > 1 - \epsilon$, then the mission is evaluated as completed, and a stop signal is sent to the planner. 

\subsubsection{Frontier Detection and Exploration}
If the probability of the object of interest $o_I$ being in the target object category $c_T$ is smaller than a custom threshold $\tau$, i.e., $ p_I(c_T) \leq \tau$, the intermediate goal is to perform exploration at likely target position. The robot's target $F$ is the frontier region $\mathbb{F}_t$. The Frontier region $\mathbb{F}_t$ is the set of cells between free and unknown space in the grid map $\mathbf{G}_t$. Formally, a grid cell $(i, j)$ belongs to the frontier region if and only if $\mathbf{G}_t[i, j] = 0$ and $
    \exists k \in \{0, 1, -1\},  \exists l \in \{0, 1, -1\}\colon \mathbf{G}_t[i+k, j+l] = -1$.

\begin{figure}
    \centering
    \includegraphics[width =0.6\linewidth]{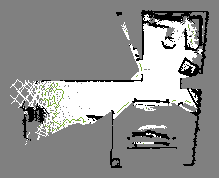}
    \caption{An example frontier region computed at time $t$, marked in green.}
    \label{fig:frontier region}
\end{figure}

Our system uses the Canny edge detector \cite{canny1986computational} to detect the grid cells between free and unknown. The detected cells are grouped into edges using 8-connectivity, i.e., each cell with coordinates $(i\pm1,j\pm1)$ is connected to the cell at $(i,j)$. Similar to map objects, a frontier edge $e_j$ is also assigned a single room ID $r_j$ based on its centroid position. The frontier region is defined as $\mathbb{F}_t = \{\langle e_1, r_1 \rangle, \langle e_2, r_2\rangle, \ldots, \langle e_m, r_m \rangle \}$, where $m$ is the number of frontier edges. Edges with area $\lvert e_j \rvert$ smaller than 
%15 
$\delta$ cells are deemed as noise and excluded from $\mathbb{F}_t$. 
% As will be discussed in more detail in section \ref{sec: reward}, the robot is encouraged to explore the frontier edge where the target object is predicted to be. 
Fig. \ref{fig:frontier region} shows an example frontier region in green.

The reward function $R(s, a, s')$ is designed as:
\begin{equation}
    R(:, :, s') = P(\mathbf{x} \in e_j \mid s') \cdot P(c_T \mid \mathbb{E}, r_i) \cdot \lvert e_j \rvert,
    \label{eq: F reward}
\end{equation}
where $P(\mathbf{x} \in e_j \mid s')$ is the probability of the robot being at frontier edge $e_j$ if its mean position is $s'$. $P(c_T \mid \mathbb{E}, r_i)$ is the probability to find target object instance $c_T$ in geometric room $r_i$ where edge $e_j$ lies given the current evidence $\mathbb{E}$. 

By including the $P(c_T \mid \mathbb{E}, r_i)$ term in the reward function, the robot is encouraged to explore frontier edges where the target object is more likely to be found. $\lvert e_j \rvert$ is the size of the frontier edge, representing the possible information gain by exploring $e_j$. $P(\mathbf{x} \in e_j \mid s')$ can be calculated by first discretizing the robot's Gaussian position distribution (with mean at $s'$) based on $\mathbf{G}_t$ and then 
summing up the probability of the robot at each cell that belongs to $e_j$. $P(c_T \mid \mathbb{E}, r_i)$ is calculated using the Bayesian network, as discussed in Section \ref{sec: semantic prior}.

\subsubsection{Object Uncertainty Reduction}
\begin{figure}
    \centering
    \includegraphics[width =0.6\linewidth]{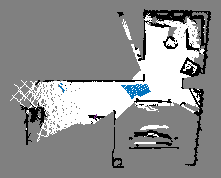}
    \caption{An example visibility region computed for one object instance at time $t$, marked in blue.}
    \label{fig:visibility region}
\end{figure}

If the probability of the object of interest $o_I$ being in the target object category $c_T$ is smaller than $1 - \epsilon$, but greater than $\tau$, i.e., $\tau < p_I(c_T) \leq 1 - \epsilon$,  the intermediate goal is to reduce uncertainty. The robot's target $F$ is the visibility region $\mathbb{V}_t$ for the object of interest $o_I$. At time $t$, the visibility region $\mathbb{V}_t$ for $o_I$ in the grid map $G_t$ with obstacles is the region of all cells on the grid map $G_t$ that $o_I$ is visible. That is, if a line connecting the position of $o_I$ and a cell $q$ does not intersect with any obstacle cell and is within the sensing range, then $q \in \mathbb{V}_t$. We apply a uniform ray-casting algorithm to compute the visibility region. Rays originating from the object's position are cast in many directions. Regions illuminated by the ray (reached by it) are considered the visibility region $\mathbb{V}_t$. An example visibility region for the current object of interest is drawn in blue in Fig. \ref{fig:visibility region}.
 
%Formally, we can define the visibility region $\mathbb{V}_t$ as such. Let $p$ be the grid corresponding to the position of $o_I$ in the grid map $\mathbf{G}_t$. Then, the visibility region $\mathbb{V}_t$ is the set of grids in the grid map $\mathbf{G}_t$, such that for every grid $q \in \mathbb{V}_t$, the line segment $pq$ does not intersect any obstacle grids. Taking sensor ability into consideration, we also require the distance $\lvert pq \rvert$ to be within the minimum and maximum sensing range $\lvert pq \rvert \in (d_{max}, d_{min})$.

The reward function $R(s, a, s')$ is designed as:
\begin{equation}
    R(:, :, s') = P(\mathbf{x} \in \mathbb{V}_t \mid s'),  
    \label{eq: v reward}
\end{equation}  
which $P(\mathbf{x} \in \mathbb{V}_t \mid s')$ is the probability of the robot being in visibility region $\mathbb{V}_t$ if its mean position is $s'$.

\section{Planning and
Execution}
We now describe how the optimal policy for reaching an intermediate goal is computed for the robot and how the robot will adapt to the ever-updated SLAM results.

\subsection{Probabilistic Planning}
The MDP $\mathbf{M}_t$ and the selected reward function $R$ are fed into a planner based on the Real Time Dynamic Programming (RTDP) algorithm \cite{smith2006focused} to compute an optimal policy $\pi^*$ that maximizes the expected sum of rewards, i.e., value function $\mathcal{V}$. A value function $\mathcal{V}$ starting at state $s \in S$ following policy $\pi$ is defined as follows:
\begin{align*}
  \mathcal{V}^\pi(s) = \operatorname{E}_{\pi}[\sum_{t = 0}^{\infty} \gamma^t R(s_t, a_t, s_{t+1})],
  %\label{eq:policy_value}
\end{align*}
where $\pi \colon S \to  A $ is a deterministic policy over $\mathbf{M}_t$ mapping the state into an action, and $\gamma \in [0, 1)$ is a discounting factor. The optimal policy $\pi^\ast$ is computed as follows: for all~$s \in S$,
\begin{align*}
  \pi^\ast(s) = \argmax_{\pi \in \Pi} \mathcal{V}^\pi(s).
  \label{eq:optimal_policy_maxprob}
\end{align*}

%Though the optimal policy can be computed by traditional dynamic programming-based approaches such as value iteration \cite{bellman1957markovian}, such approaches can not meet the real-time criteria when the state space $S$ is large. To tackle this issue, we use the Real Time Dynamic Programming (RTDP) algorithm \cite{smith2006focused}. 
The RTDP algorithm allows us to compute a semi-optimal policy in a short time\footnote{unlike a more traditional approach such as value iteration \cite{bellman1957markovian}.}. As the robot carries out the semi-optimal policy, the policy is continuously improved by the RTDP algorithm with the current robot mean position as the initial state $s_0$ and converges to the optimal policy.  

% \begin{algorithm}
% \caption{RTDP algorithm}
% \label{alg:RTDP}
% \While{more trials required}{
%    $s = s_0$\;
%    \While{$s \notin F$}{
%    $a = \argmax_{a \in A} \sum_{s'\in S} P(s, a, s')\cdot (\mathcal{V}(s') + R(s, a, s')) $\;
%    $\mathcal{V}(s) = \sum_{s'\in S} P(s, a, s')\cdot (\mathcal{V}(s') + R(s, a, s')) $\;
%    $s' \sim P(s, a, s')$
%    }
% }
% \end{algorithm}
\subsection{Adaptation}
The fused map $E_t$, the robot pose estimation $bel(\mathbf{x}_{t})$, frontier region $\mathbb{F}_t$, and visibility region $\mathbb{V}_t$ are updated at every time instance $t$ based on the ever-improving SLAM results. Consequently, once the robot reaches an intermediate target state, the MDP model $\text{M}_t$ must be updated. We call this process the {\em adaptation} of the MDP model. Next, the corresponding policy $\pi$ is also updated. %Once the state $s$ in the MDP has reached an intermediate states $F$, the MDP model would be updated.

Specifically, the following components are adapted: 
\begin{itemize} 
\item the discrete state space $S$ to match the changing grid map $\mathbf{G}_t$, 
\item the initial state $s_0$ to match the changing robot pose estimation $bel(\mathbf{x}_{t})$, 
\item the transition probability function $P$, 
\item the reward function $R$ based on Eqs. \eqref{eq: v reward} and \eqref{eq: F reward}, and 
\item the set of intermediate target states $F$ as  $\mathbb{F}_t$ and $\mathbb{V}_t$ change. 
\end{itemize}
The RTDP planner takes the updated MDP model $\text{M}_t$ to generate a new policy.

\section{EXPERIMENTS}

We conducted experiments in both simulated and real environments to verify the effectiveness and efficiency of our approach. 

\subsection{Experiment Setup}
\begin{figure}
     \centering
     \begin{subfigure}[b]{0.49\linewidth}
         \centering
\includegraphics[width=\textwidth]{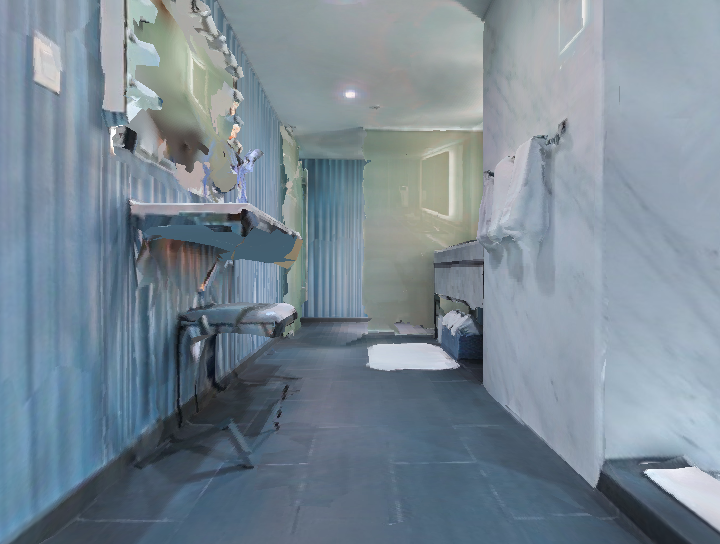}
     \end{subfigure}
     \begin{subfigure}[b]{0.49\linewidth}
         \centering
\includegraphics[width=\textwidth]{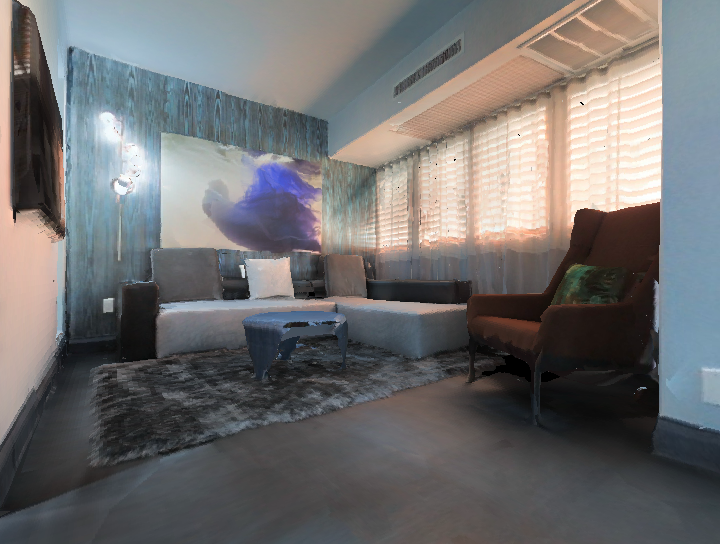}
     \end{subfigure}
        \caption{Two snap shots of the MP3D scene.}
        \label{fig: snap shots}
\end{figure}
\begin{figure}
    \centering
    \includegraphics[width=\linewidth]{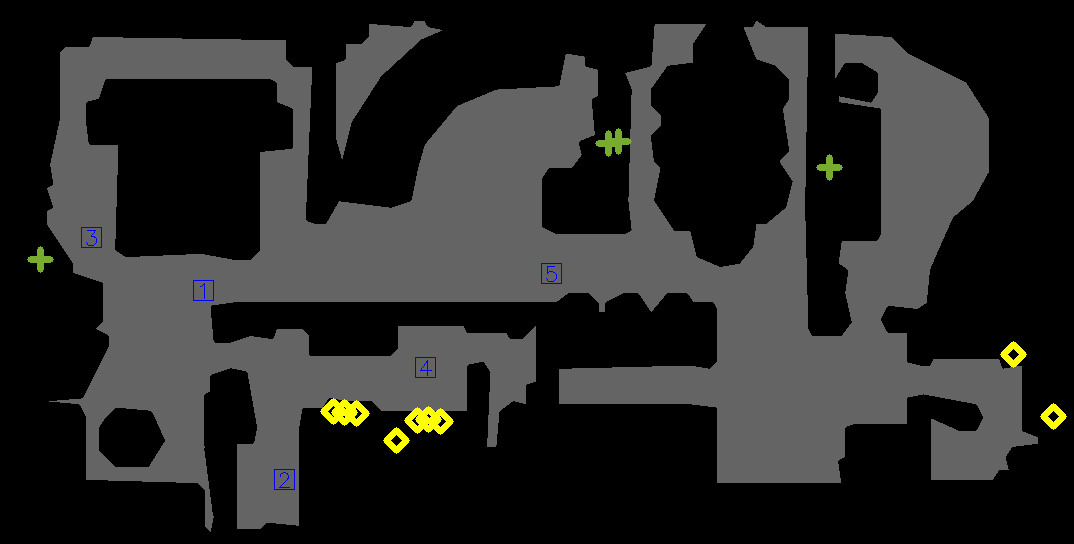}
    \caption{Target objects (yellow diamonds for towel, green cross for TV monitors) and five random robot starting positions (blue numbered boxes).}
    \label{fig:start and goal}
\end{figure}
In each experiment, the robot's objective is to find any instance of the target with a confidence level greater than $1 - \epsilon = 0.99$. The robot's linear and angular velocity actuation signal is continuous. A task is successful if the agent has identified the target object within a specific time budget (1K seconds) and stops its motion. The set of parameters used for virtual and real-world experiments are presented in Table \ref{tab:param}.

\begin{table}[htbp]
    \caption{Parameters for virtual and real-world experiments}
    \centering
    \begin{tabular}{c c c}
    \toprule
        Parameter & Parameter meaning & Value   \\
    \midrule 
 $\epsilon$ & Threshold for mission completion & 0.01   \\
  $\tau$ & Threshold for exploration & 0.5 \\
  $\lambda$ & Threshold for object evidence & 0.5 \\
 $\gamma$ & discount factor of the MDP & 0.9 \\
 $\delta$ & Threshold for frontier region filter & 15 \\
    \bottomrule
    \end{tabular}
    \label{tab:param}
\end{table}

%The accompanying video shows the robot's operations to find target objects in simulation and real environments. 
\subsection{Virtual-World Study}
We performed experiments on the Matterport3D (MP3D) \cite{chang2017matterport3d} dataset using the Habitat \cite{savva2019habitat} simulator. MP3D dataset contains 3D scenes of a typical indoor environment, and the Habitat simulator allows the robot to navigate the virtual 3D scene. Two snapshots of the MP3D scene are given in Fig. \ref{fig: snap shots}. This particular scene is $6.4m\times 8.4m$ in size and has one level, ten rooms, and 187 objects, but it is unknown to the robot. The robot start positions and the target object instances are indicated in Fig. \ref{fig:start and goal}. 

\begin{figure}[htbp]
     \centering
     \begin{subfigure}[b]{1\linewidth}
         \centering
         \includegraphics[width=\textwidth]{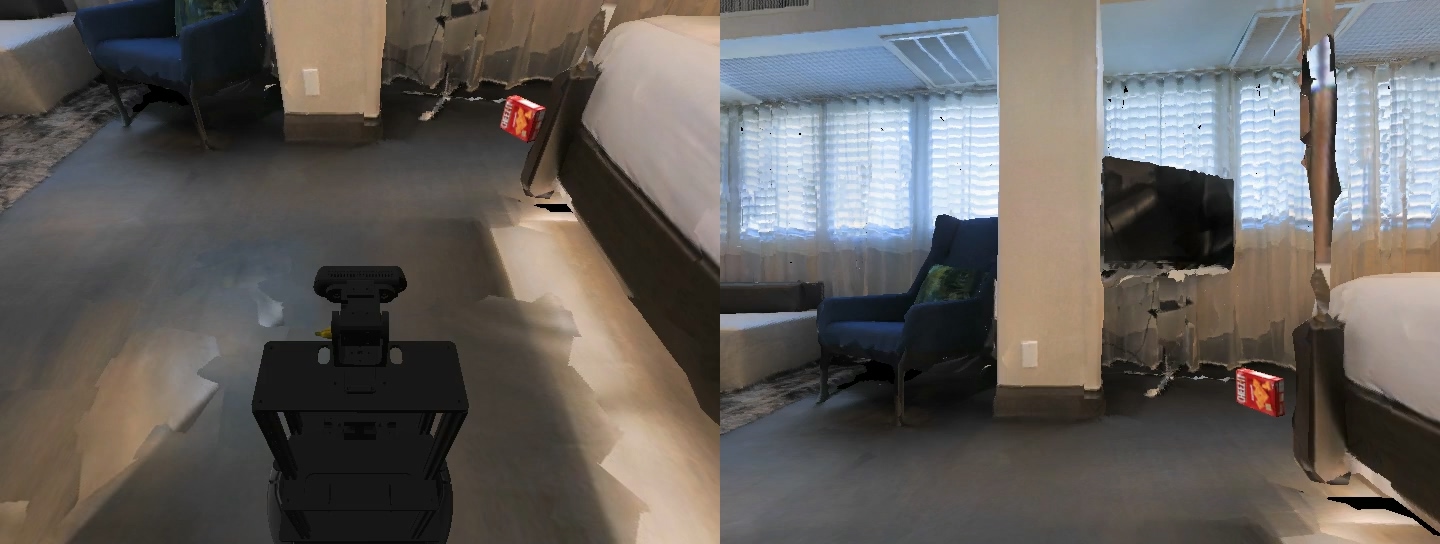}
         \caption{View at the position 1, the initial position }
     \end{subfigure}
     \begin{subfigure}[b]{1\linewidth}
         \centering
         \includegraphics[width=\textwidth]{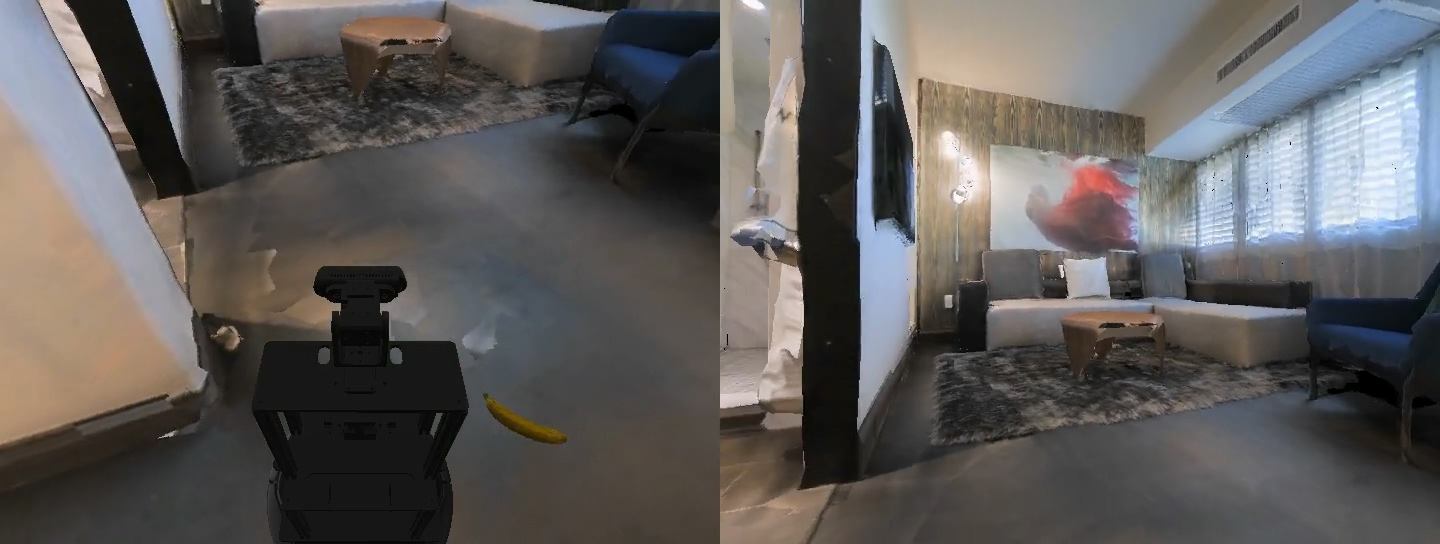}
         \caption{View during exploration task}
     \end{subfigure}
      \begin{subfigure}[b]{1\linewidth}
         \centering
         \includegraphics[width=\textwidth]{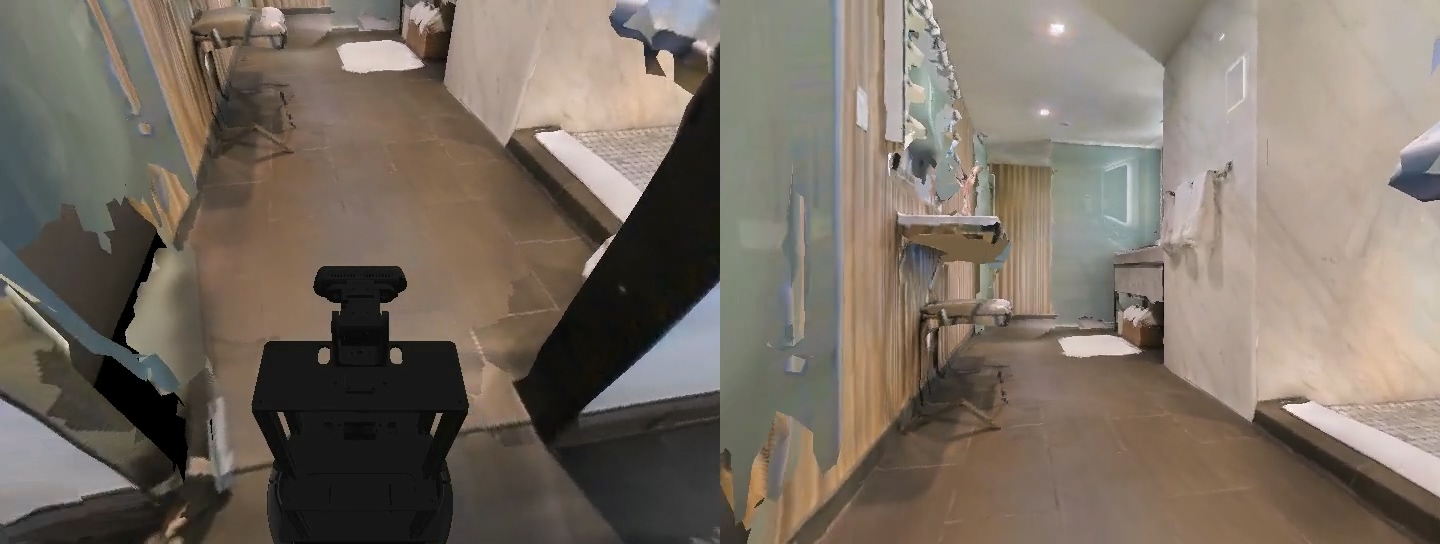}
          \caption{View during uncertainty reduction task}
     \end{subfigure}
     \begin{subfigure}[b]{1\linewidth}
         \centering
         \includegraphics[width=\textwidth]{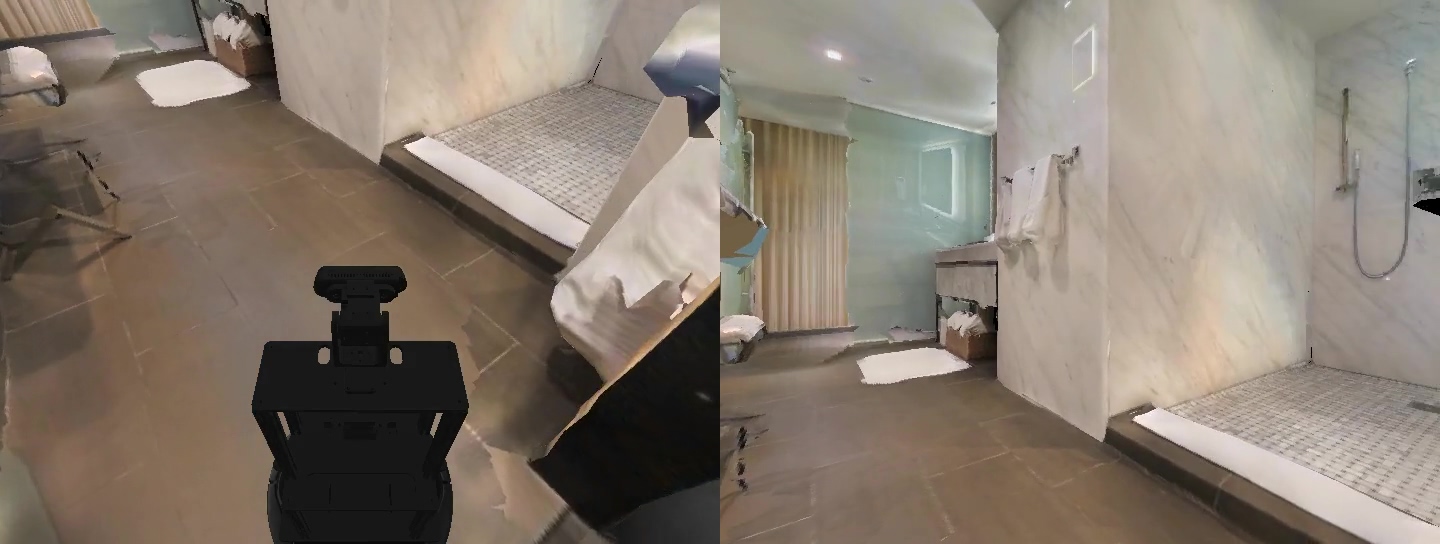}
          \caption{View at task completion when towels are found \newline}
     \end{subfigure}
        \caption{Third-person view (left) and first-person view of the robot (right) during task execution to find towels.}
        \label{fig: virtual world demo}
\end{figure}

Random noise following a Dirichlet distribution is injected into the ground truth semantic labels that the Habitat-lab simulator provides to simulate the results from an object detector. Random Gaussian noise is also injected into the linear and angular velocity of the mobile robot to simulate the motion uncertainty. We used an 
AMD Ryzen\textsuperscript{TM} 3700X CPU with 3.60GHz, 16 GB of RAM, and an Nvidia GeForce RTX 3060Ti graphics card for all simulations. Fig. \ref{fig: virtual world demo} presents several snapshots showing the robot starting from position 1 in the virtual environment at various stages of the task of finding towels. 

\subsubsection{Semantic SLAM results}
\begin{figure}[htbp]
\centering\includegraphics[width=0.79\linewidth]{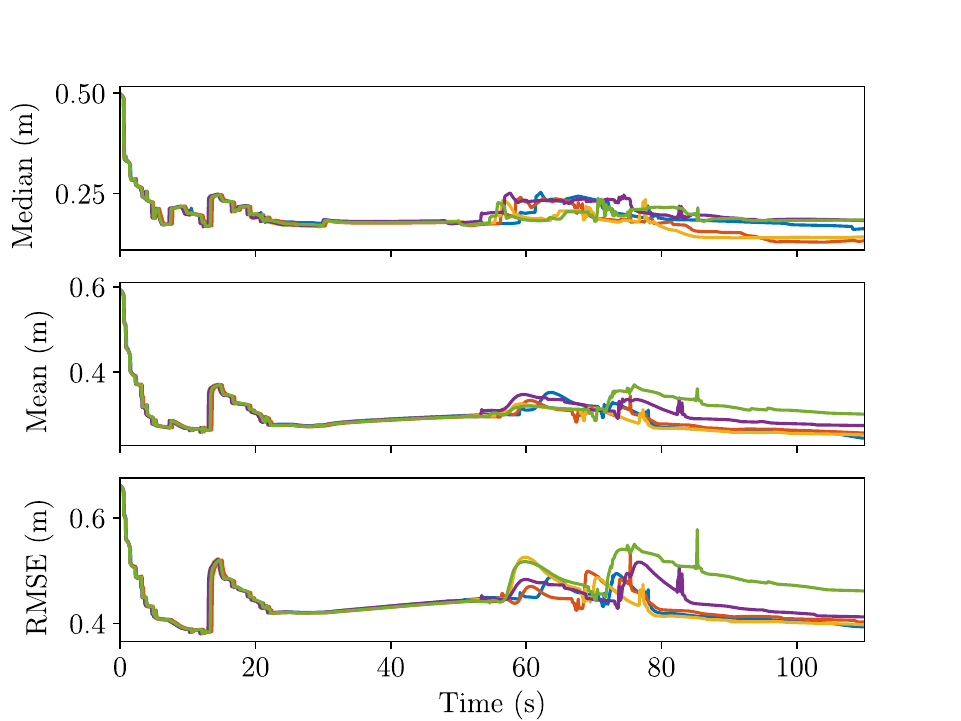}
    \caption{Median, Mean and RMSE of the error between the predicted and ground-truth positions of all the objects detected. Each colored curve represents a separate run. }
    \label{fig: position error}
\end{figure}

\begin{figure}[htbp]
\centering\includegraphics[width=0.79\linewidth]{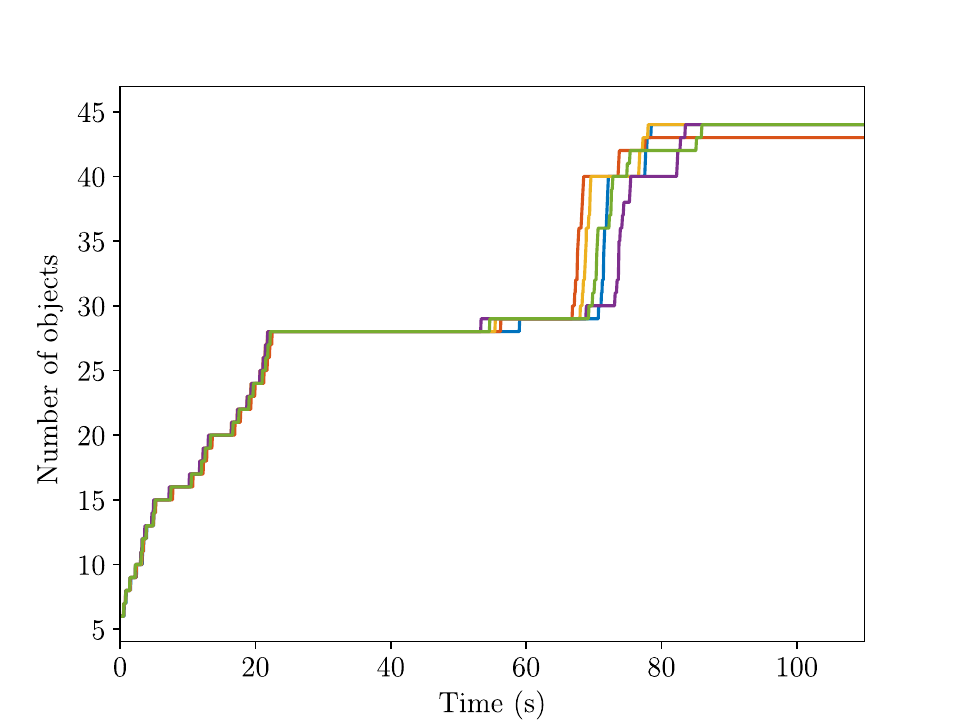}
    \caption{Number of identified objects over time. Each colored curve represents a separate run.}
    \label{fig: obj num}
\end{figure}

\begin{figure}[htbp]
    \centering\includegraphics[width=0.79\linewidth]{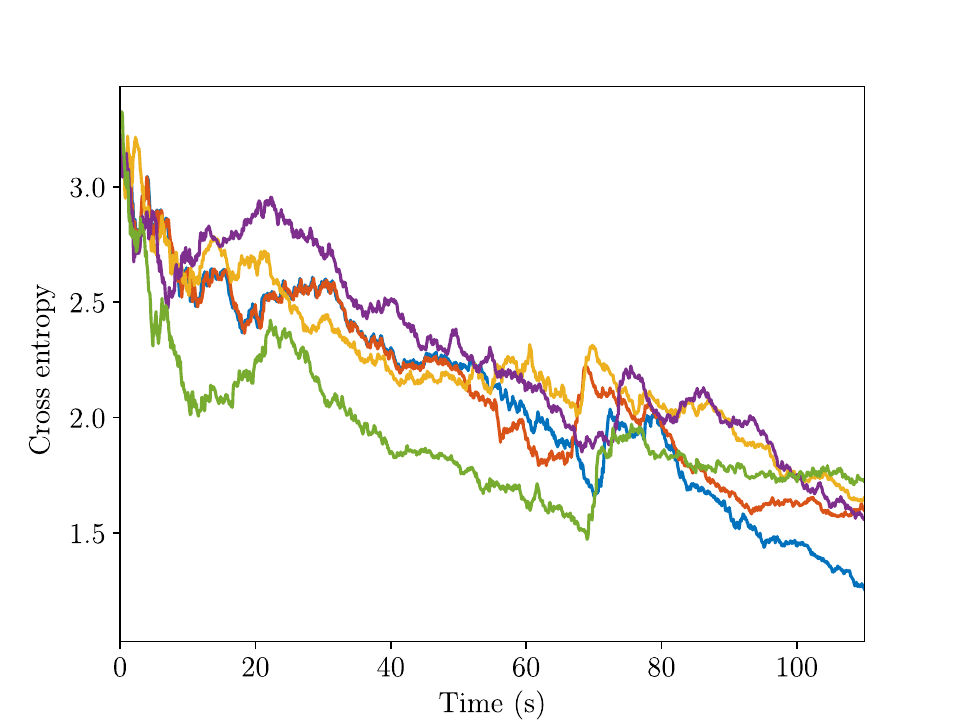}
    \caption{Cross entropy between the predicted and the ground-truth classes (available from Habitat Simulator) of all the objects detected. Each colored curve represents a separate run.}
    \label{fig: cross entropy}
\end{figure}

\begin{figure}
    \centering
    \includegraphics[width=1\linewidth]{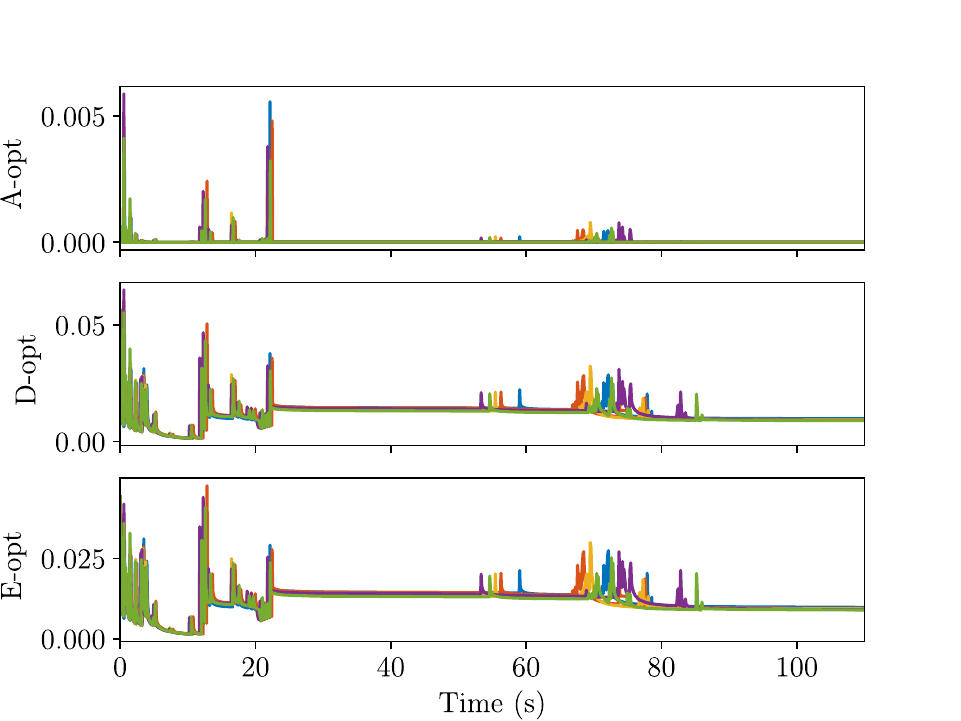}
    \caption{The evolution of position covariance of all the objects detected over time. Each colored curve represents a separate run.}
    \label{fig: obj covariance}
\end{figure}

\begin{figure}
     \centering
     \begin{subfigure}[b]{1\linewidth}
         \centering
         \includegraphics[width=1\textwidth]{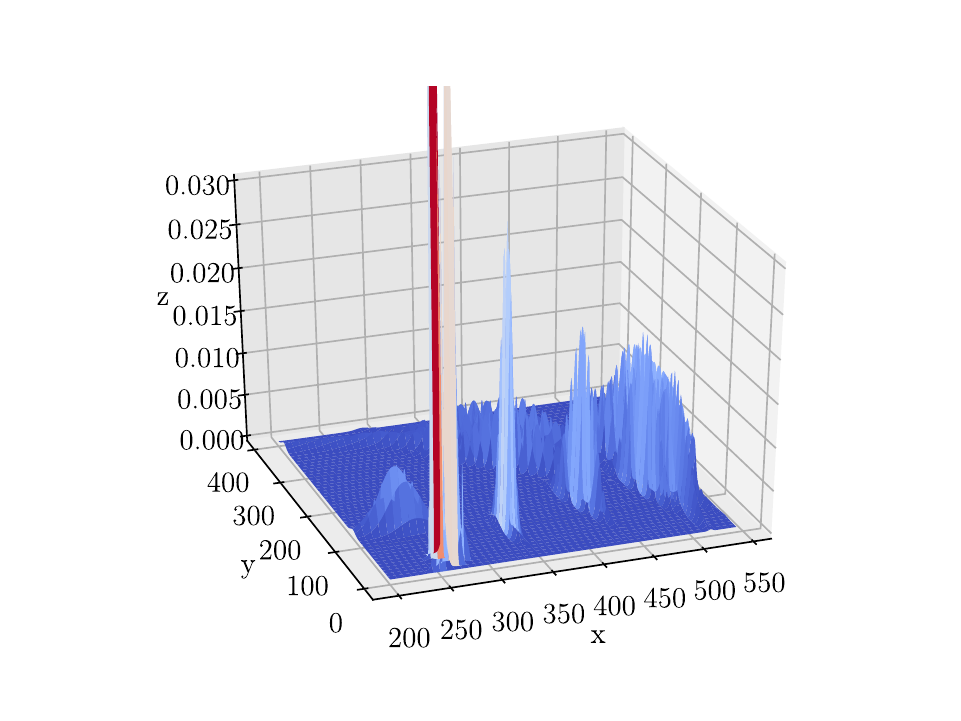}
         \caption{$t=8s$.}
     \end{subfigure}
     \begin{subfigure}[b]{1\linewidth}
         \centering
         \includegraphics[width=1\textwidth]{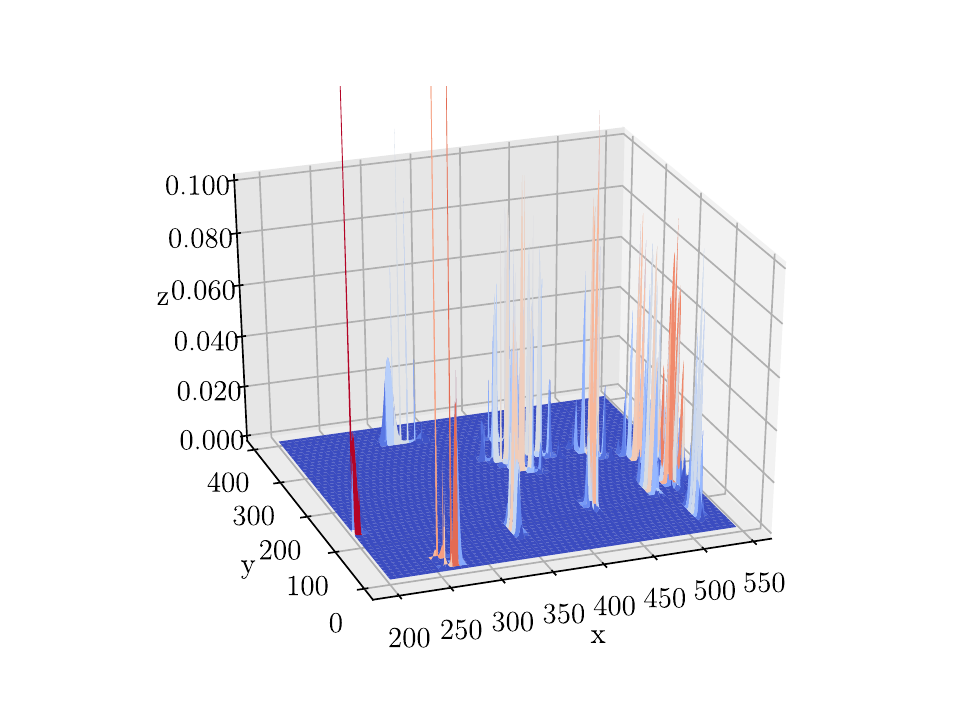}
         \caption{$t=70s$.}
     \end{subfigure}
        \caption{Object position covariance at two time instants}
        \label{fig: covariance map}
\end{figure}

\begin{figure}
    \centering
    \includegraphics[width=0.79\linewidth]{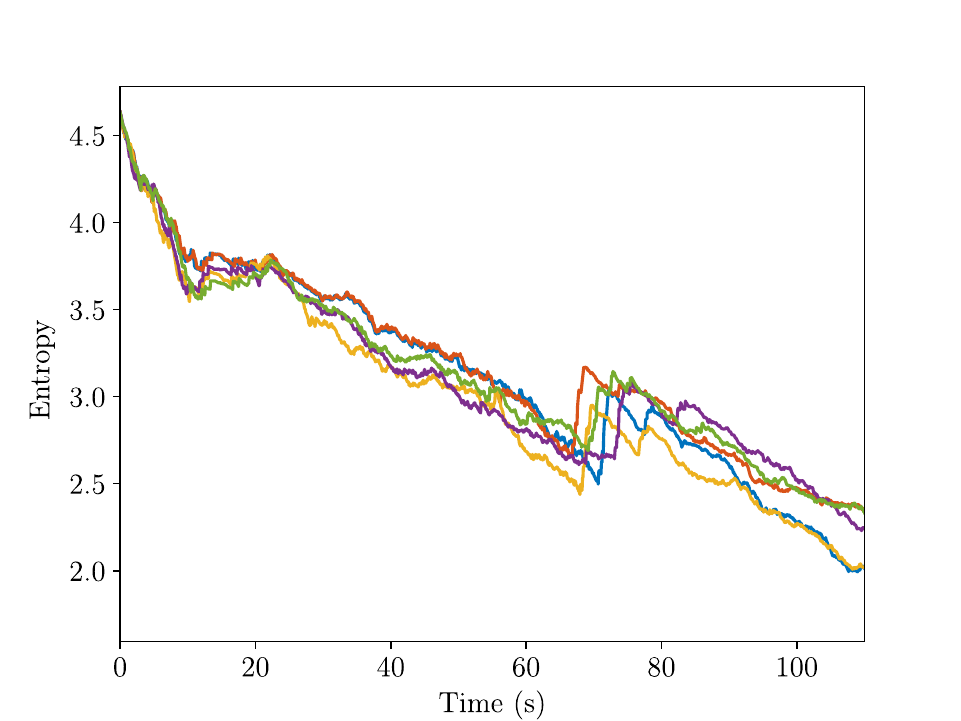}
    \caption{The evolution of the predicted class entropy of all detected objects over time. Each colored curve represents a separate run.}
    \label{fig: entropy}
\end{figure}

We present the semantic SLAM results obtained by running five task runs from starting position 1 in the MP3D dataset. Our evaluation focuses on the accuracy and uncertainty of the collected results. The results are summarized in Table \ref{tab:slam results}, visualized in Figs. \ref{fig: position error} -- \ref{fig: entropy}, and explained in details below. 

\begin{table}[htbp]
    \caption{Semantic SLAM results tested on the MP3D dataset from five separate runs (T1--T5).}
    \centering
    \begin{tabular}{c c c c c c c}
    \toprule
        Metrics & T1 & T2 & T3 & T4 & T5 & Average\\
    \midrule 
    MEDIAN(m)     &  0.16 & 0.14 & 0.15 & 0.17 & 0.18 & 0.26 \\
    MEAN(m)     &  0.25 & 0.25 & 0.25 & 0.27 & 0.31 & 0.27 \\
    RMSE(m)     &  0.39 & 0.39 & 0.39 & 0.43 & 0.48 & 0.42 \\
    Cross Entropy &  1.28 & 1.31 & 1.37 & 1.15 & 1.27 & 1.27 \\
    Entropy &  1.93 & 1.80 & 1.73 & 1.75 & 1.86 & 1.82 \\
    A-OPT ($e^{-03}$) & 9.8 & 9.3 & 9.3 & 9.0 & 14.7 & 10.4 \\
    D-OPT ($e^{-06}$) &  6.95 & 6.28 & 6.29 & 7.02 & 45.7 & 14.5 \\
    E-OPT ($e^{-03}$) &  9.7 & 9.3 & 9.2 & 9.0 & 14.5 & 10.3 \\
    \bottomrule
    \end{tabular}
    \label{tab:slam results}
\end{table}

\paragraph{Accuracy}
We calculate the mean and the median of the error between a predicted object's position and the ground-truth object's position:
\begin{align*}
\operatorname {Mean} & =\textstyle \sum_{i=1}^{n}\lVert{\hat {p}}_{i}-p_{i}\rVert / n,\\
\operatorname {Median} & = \operatorname {median}(\lVert{\hat {p}}_{i}-p_{i}\rVert),\\
\operatorname {RMSE} & =\sqrt{\textstyle \sum_{i=1}^{n}\lVert{\hat {p}}_{i}-p_{i}\rVert^2 / n} 
\end{align*}
% \begin{equation*}
% \text{Mean} =\frac {\sum _{i=1}^{n}\lVert{\hat {p}}_{i}-p_{i}\rVert}{n}, \hspace{2mm} \text{Median} = \operatorname {median}(\lVert{\hat {p}}_{i}-p_{i}\rVert),
% \end{equation*}
\noindent where $n$ is the current number of objects, $\hat {p}_i$ is the estimated object position, and $p_i$ is the ground truth object position. Their variations over time are plotted in Fig. \ref{fig: position error}, and the final error is presented in Table \ref{tab:slam results}.  For reference, the number of identified objects at each time instance is also plotted in Fig. \ref{fig: obj num}. We can see that the error increases for the first few seconds as new objects are identified. Nonetheless, as more observations come in, the error decreases and converges. The final error in Table \ref{tab:slam results} is partially due to the difference between the center of object as defined by our semantic SLAM system and the ground truth. For the same reason, we expect large errors for objects with large sizes, such as tables or sofas, and we consider the median error as a more sensible metric.

We also calculate the cross entropy between the classes of the predicted objects and those of ground-truth objects: 
\begin{equation*}
    -\frac{1}{n}\sum_{i=1}^{n}\sum _{c\in {\mathbb {C}}}p^{gt}_i(c)\,\log p_i(c),
\end{equation*} 
$p_i(\cdot)$ is the predicted object class distribution, $p_i^{gt}(\cdot)$ is the ground truth of object class distribution, taking the value one at the corresponding object class and zero elsewhere. The results are plotted in Fig. \ref{fig: cross entropy}, and the final cross-entropy is presented in Table \ref{tab:slam results}. We can see that the cross entropy gradually decreases with time and reaches a small value in the end, proving that the classes of the predicted objects will converge to the correct results. 

\paragraph{Uncertainty}
%Though we do not claim our method to be an active SLAM method, 
We observe a decrease in semantic map uncertainty as the robot progresses. The average A-opt (sum of the covariance matrix eigenvalues), D-opt (product of covariance matrix eigenvalues), and E-opt (largest covariance matrix eigenvalue) of the map object position covariance are calculated. Their evolution over time is plotted in Fig. \ref{fig: obj covariance}, and the final values are stored in Table \ref{tab:slam results}. The spikes in the graph indicate new objects are identified, hence the increased position uncertainty. However, as time passes and more observations come in, we can see that all three metrics are kept at a low level.  This shows that the robot can estimate the objects' position fairly confidently.

Fig. \ref{fig: covariance map} gives a more intuitive representation. In Fig. \ref{fig: covariance map}, we plot the Gaussian functions with their means and covariances set as the estimated object position means and covariances. Therefore, each ``bell'' in the plot represents one object. Comparing the results we obtained at time instants $t=8s$ and $t=70s$, we can see that at $t=70s$, the bell's peak increases, and the base decreases, indicating a more accurate estimation of the object's position.  

The entropy of the predicted object class is also calculated: 
\begin{equation*}
    -\frac{1}{n}\sum_{i=1}^{n}\sum _{c\in {\mathbb {C}}}p_i(c)\,\log p_i(c)
\end{equation*} 
and visualized in Fig. \ref{fig: entropy}, its final value is also listed in Table \ref{tab:slam results}. The result suggests that as time progresses, the robot is more and more certain about the object class it predicted. 

\subsubsection{Planning results}

We use the following metrics to evaluate a planner's performance: 
\begin{itemize}
\item {\bf Success}: percentage of successful task runs.
\item  {\bf Average path length}: average length of the path taken by the agent in all task runs.
\item  {\bf Success weighted by path length (SPL)} \cite{anderson2018evaluation}: 
% $\frac{1}{N}\sum_{i=1}^{N}S_i \frac{l_i}{\max(p_i, l_i)}$, 
    \begin{equation*}
       \operatorname{SPL} = \frac{1}{N}\textstyle \sum_{i=1}^{N}S_i \frac{l_i}{\max(p_i, l_i)},
   \end{equation*}
where $l_i$$=$length of the shortest path between goal and the visibility region of target instance for a run, $p_i$$=$length of the path taken by the agent in a run, $S_i$$=$ binary indicator of success in the $i$th run.
\item  {\bf Average time consumption}: average time spent on a single task run.
\end{itemize}

%\noindent {\bf Success}: percentage of successful task episodes\\
%\noindent {\bf Average path length}: average length of the path taken by the agent in an episode.\\
%\noindent {\bf Success weighted by path length (SPL)} \cite{anderson2018evaluation}: 
%$\frac{1}{N}\sum_{i=1}^{N}S_i \frac{l_i}{\max(p_i, l_i)}$, 
    %\begin{equation*}
      %  \operatorname{SPL} = \frac{1}{N}\sum_{i=1}^{N}S_i \frac{l_i}{\max(p_i, l_i)},
   % \end{equation*}
%\noindent where $l_i$$=$length of the shortest path between goal and the visibility region of target instance for an episode, $p_i$$=$length of the path taken by the agent in an episode, $S_i$$=$ binary indicator of success in episode $i$.\\
%\noindent {\bf Average time consumption}: average time spent on a single episode.

Further, to establish the effectiveness of the individual components used in our planner, we conducted  the following ablation studies: 
\begin{itemize}
 \item  \textbf{Method-S}: ablation of our method without semantic prior knowledge and using a uniform reward.
\item \textbf{Method-UR}: ablation of our method without uncertainty reduction.  
\end{itemize}

We tested all methods on two target object categories: Towel and TV Monitor. Our system sampled five random start positions, and from each start position, it conducted five task runs. The average results are shown in Table \ref{tab:average}. 

Our method outperforms the ablation baselines in success rate, SPL, and path length by a large margin. It in turn proves the efficacy of using semantic prior knowledge to guide the search for the target object and the necessity of taking explicit actions to reduce map uncertainty. 

\begin{table}[htbp]
    \caption{Comparison study}
    \centering
    \begin{tabular}{c c c c c c}
    \toprule
        Category & Method & Success & Path length (m) & SPL & Time (s)\\
    \midrule 
  \multirow{3}{4em}{Towel} &  Our Method     & \textbf{96\%} & \textbf{4.98} & \textbf{0.39} & \textbf{242.3} \\
    & Method-S & 68\% & 8.73 & 0.28 & 291.7 \\
    & Method-UR & 92\% & 8.38 & 0.26 & 327.4 \\
    \midrule 
    \multirow{3}{4em}{TV Monitor}  &  Our Method     & \textbf{100\%} & \textbf{5.75} & \textbf{0.55} & \textbf{251.7} \\
    & Method-S & 84\% & 10.85 & 0.37 & 397.5 \\
    & Method-UR & 92\% & 9.19 & 0.39& 387.2 \\
    \bottomrule
    \end{tabular}
    \label{tab:average}
\end{table}

\subsection{Real-World Demos}
\begin{figure}[htbp]
     \centering
     \begin{subfigure}[b]{0.49\linewidth}
         \centering
         \includegraphics[width=\textwidth]{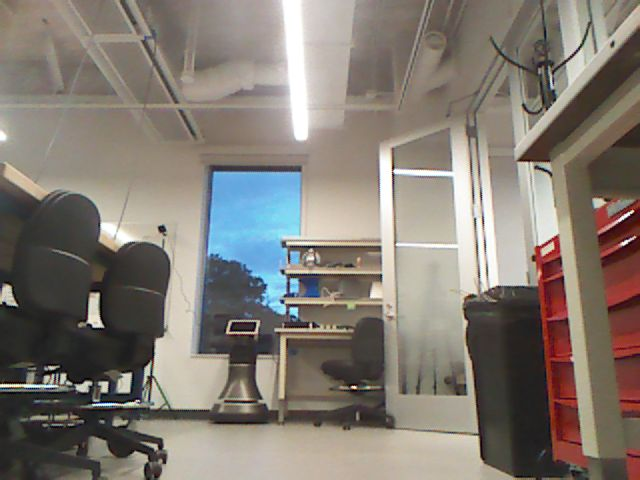}
         \caption{View at initial position}
     \end{subfigure}
     \begin{subfigure}[b]{0.49\linewidth}
         \centering
         \includegraphics[width=\textwidth]{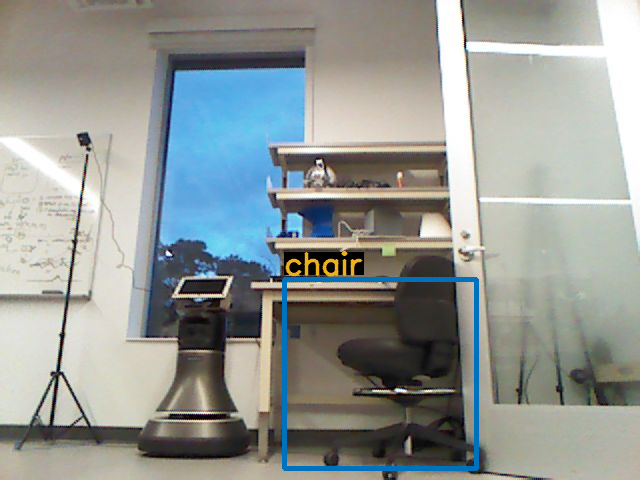}
         \caption{View during exploration}
     \end{subfigure}
          \begin{subfigure}[b]{0.49\linewidth}
         \centering
         \includegraphics[width=\textwidth]{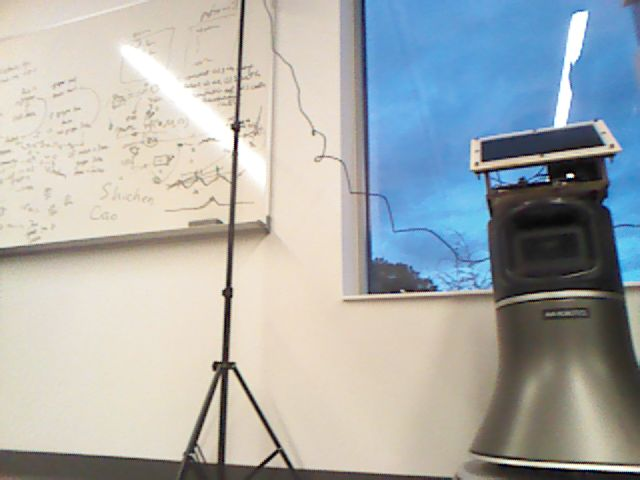}
         \caption{View during exploration}
     \end{subfigure}
     \begin{subfigure}[b]{0.49\linewidth}
         \centering
         \includegraphics[width=\textwidth]{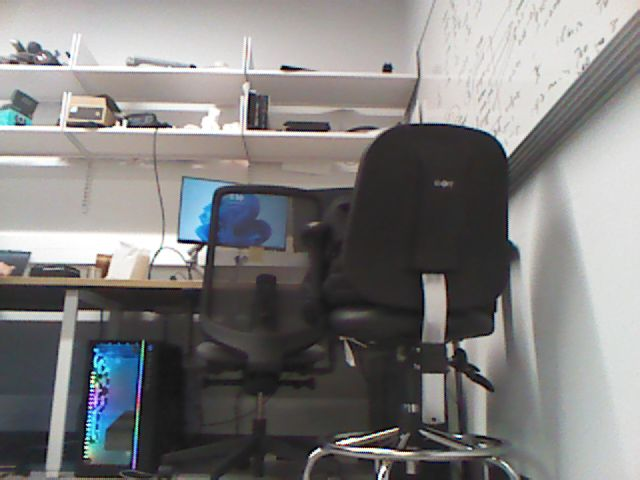}
         \caption{View during exploration}
     \end{subfigure}
      \begin{subfigure}[b]{0.49\linewidth}
         \centering
         \includegraphics[width=\textwidth]{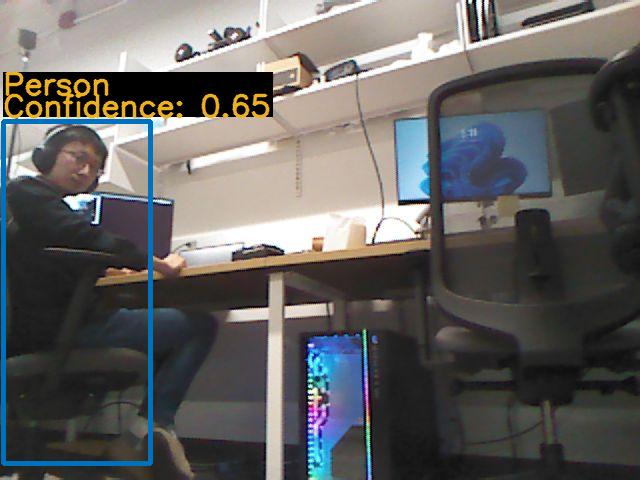}
          \caption{View during uncertainty reduction}
     \end{subfigure}
     \begin{subfigure}[b]{0.49\linewidth}
         \centering
         \includegraphics[width=\textwidth]{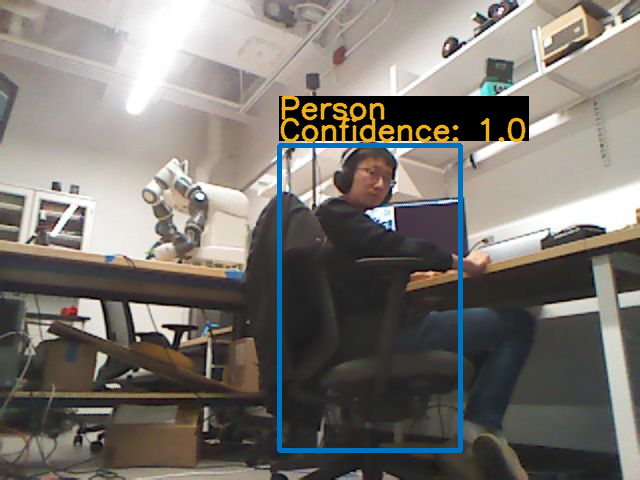}
          \caption{View at task completion \newline}
     \end{subfigure}
        \caption{First-person views of the robot during task execution to find a person in the laboratory.}
        \label{fig: real world demo}
\end{figure}

We also performed real-world experiments using a TurtleBot2 mobile robot. We deployed our algorithm on the TurtleBot and tested it in two unknown environments -- involving a cluttered laboratory and a kitchen. 

\begin{figure}[tbp]
    \centering
    \includegraphics[width=0.97\linewidth]{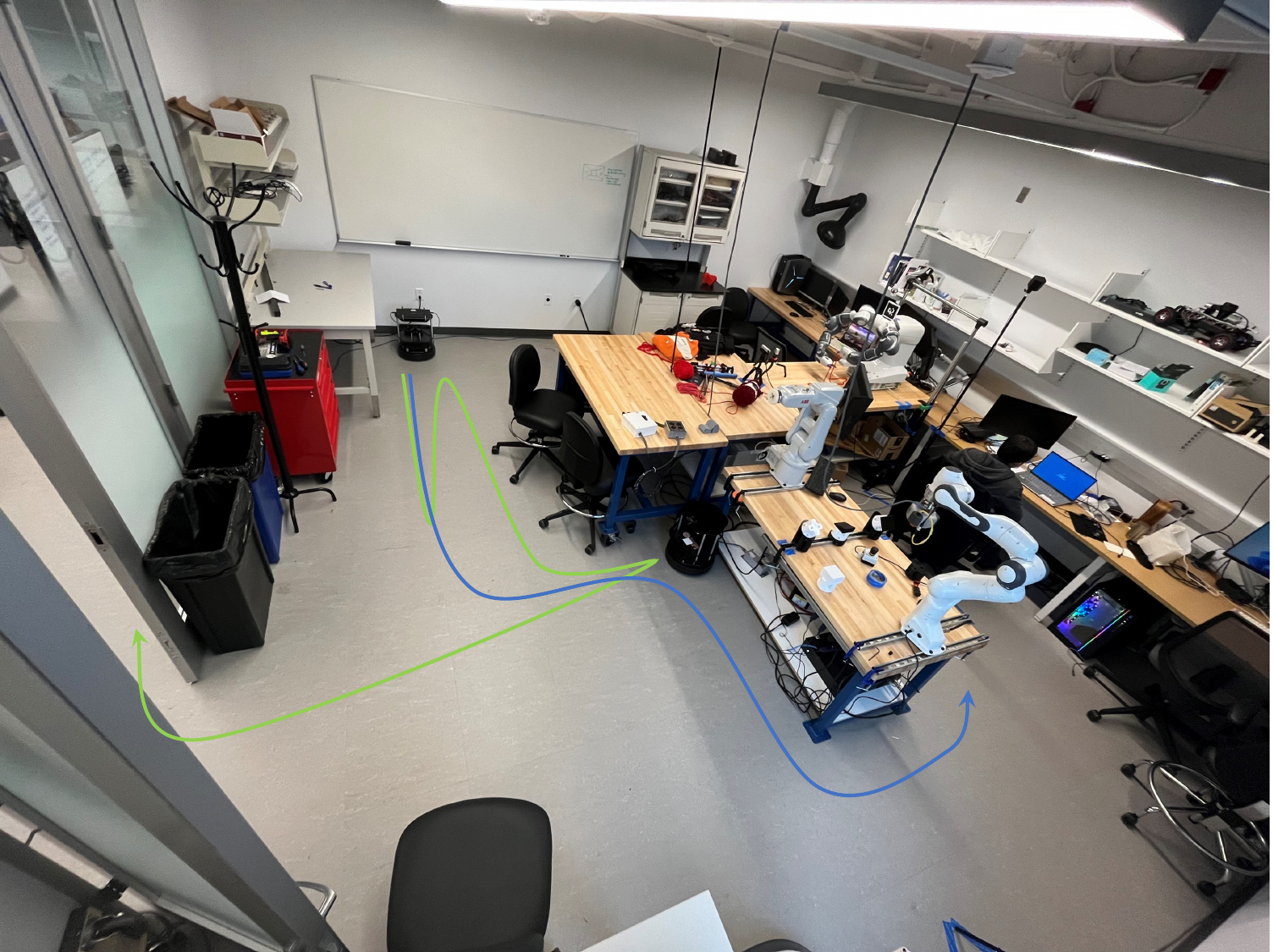}
    \caption{Setup for the real-world demonstration in a lab. 
    %The TurtleBot's initial position and the target object (person) are represented by the actual identities. 
    The success path is visualized by the blue line with semantic prior knowledge. The failure path without semantic prior knowledge is visualized by the green line.}
    \label{fig: lab setup}
\end{figure}

In the laboratory, the robot's task is to find a person. The laboratory is 6m × 6m in size, unknown to the robot. Several sample images from the laboratory environment were taken from the robot's first-person view, showing the robot at the initial position, performing exploration tasks, performing uncertainty reduction tasks, and finally identifying the target object, as shown in Fig. \ref{fig: real world demo}. In Fig. \ref{fig: real world demo}b, the robot has detected a chair in this room and inferred that a person is likely to be in the same room. From Figs. \ref{fig: real world demo}e and \ref{fig: real world demo}f, we can see that before the uncertainty reduction operation, the confidence score for the person was 0.65. After the operation, the confidence score was 1.00, and the robot correctly determined that the task was completed. The success path the robot traversed is visualized in the blue line in Fig. \ref{fig: lab setup}. 

We performed three runs in the laboratory, and all were successful. The average total time for the robot to complete the task was 
%$196$ seconds
about 4.5 minutes, and the average time spent on planning alone was $67$ seconds. The average length of the robot path was 16.3m. We estimate the optimal path to be 6.5m, which gives a SPL of 0.41. 

To highlight the importance of using semantic prior knowledge, we conducted another test by denying the robot planner access to semantic prior knowledge and using a uniform reward. Without semantic prior knowledge,  the robot chose to exit the laboratory and failed to find the target, as visualized in the green line in Fig. \ref{fig: lab setup}. 

\begin{figure}
     \centering
     \begin{subfigure}[b]{0.49\linewidth}
         \centering
         \includegraphics[width=\textwidth]{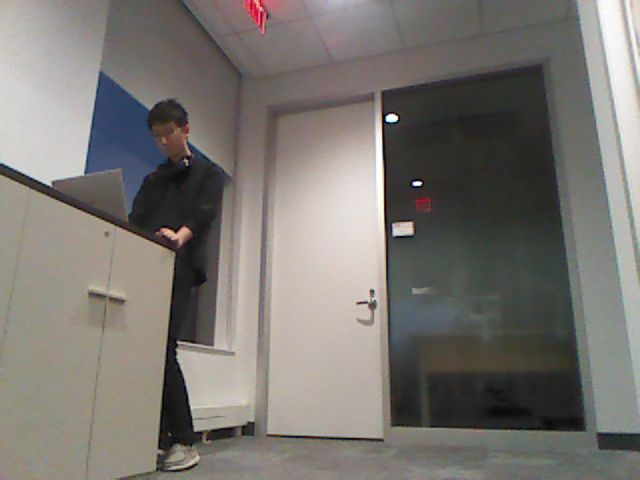}
         \caption{View at initial position}
     \end{subfigure}
     \begin{subfigure}[b]{0.49\linewidth}
         \centering
         \includegraphics[width=\textwidth]{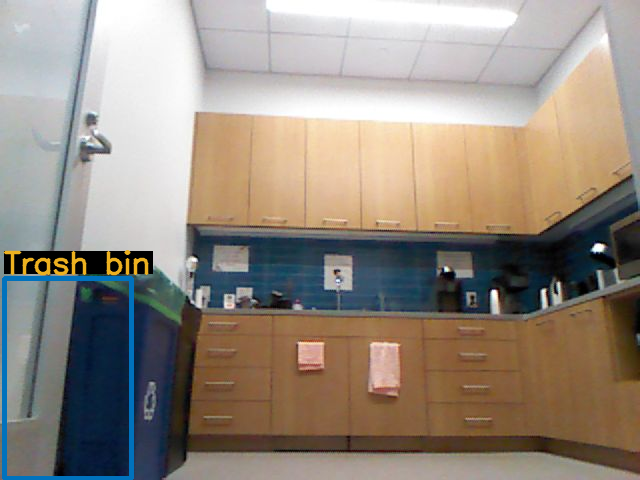}
         \caption{View during exploration}
     \end{subfigure}
      \begin{subfigure}[b]{0.49\linewidth}
         \centering
         \includegraphics[width=\textwidth]{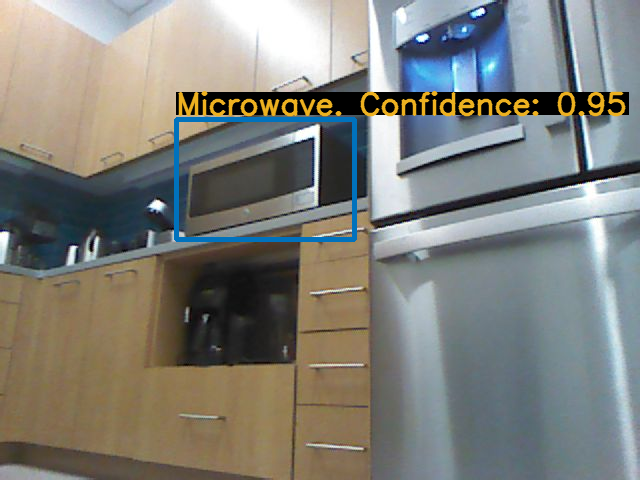}
          \caption{View during uncertainty reduction}
     \end{subfigure}
     \begin{subfigure}[b]{0.49\linewidth}
         \centering
         \includegraphics[width=\textwidth]{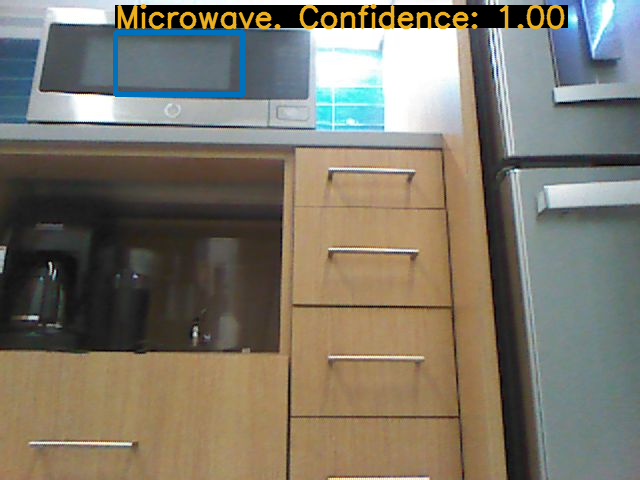}
          \caption{View at task completion \newline}
     \end{subfigure}
        \caption{First-person views of the robot during task execution to find a microwave in the kitchen scenario.}
        \label{fig: real world kitchen demo}
\end{figure}

\begin{figure}[tbp]
   \centering
    \includegraphics[width=0.97\linewidth]{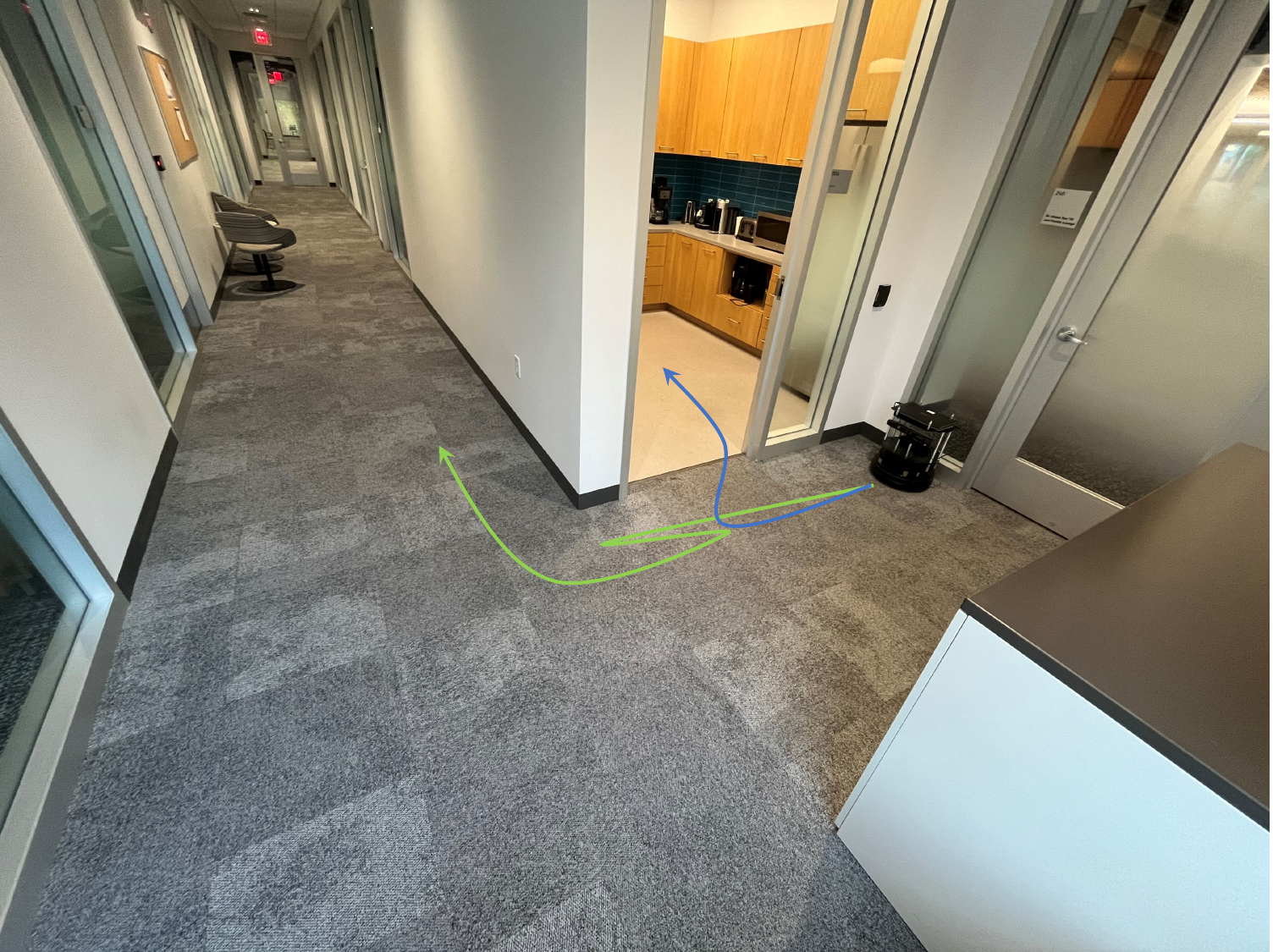}
    \caption{Setup for the real-world demonstration in a kitchen scenario. 
    %The TurtleBot's initial position and the target object (microwave) are represented by the actual identities. 
    The success path with semantic prior knowledge is visualized by the blue line. The failure path without semantic prior knowledge is visualized by the green line.}
    \label{fig: kitchen setup}
\end{figure}

\begin{figure}[htbp]
     \centering
      \includegraphics[width=0.97\linewidth]{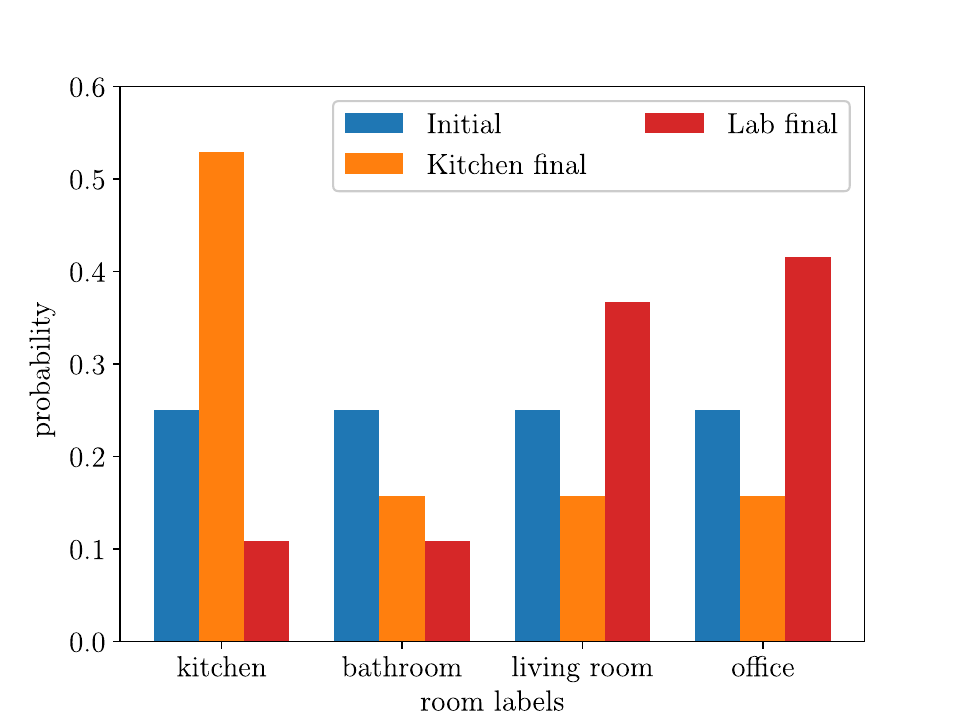}
     
        \caption{Semantic room labels inferred by the Bayesian networks at the beginning and end of the task in the lab and kitchen demos.}
        \label{fig: semantic room labels}
\end{figure}

For the kitchen scenario, the robot's task is to find a microwave. The space is about 4.5m × 5.5m in size, unknown to the robot. Several robot first-person view images from the environment are presented in Fig. \ref{fig: real world kitchen demo}. In Fig. \ref{fig: real world kitchen demo}b, the robot has detected a trash bin and inferred that the unexplored room is a kitchen and probably contains a microwave. From Figs. \ref{fig: real world kitchen demo}c and \ref{fig: real world kitchen demo}d, we can see that after the uncertainty reduction operation, the confidence score for the microwave increased from 0.95 to 1.00, and the robot correctly determined that the task was completed. The success path the robot traversed was visualized in the blue line in Fig. \ref{fig: kitchen setup}.

Again, we performed three runs in this kitchen scenario, and all three were successful. The average total time for the robot to complete the task was 7.1 minutes, and the average time spent on planning was $4.3$ minutes. The average length of the robot path is 13.0m. We estimate the optimal path to be 3m, which gives a SPL of 0.24. 

\begin{figure}[tbp]
     \centering
     \begin{subfigure}[b]{0.67\linewidth}
         \centering
         \includegraphics[height=2.7cm]{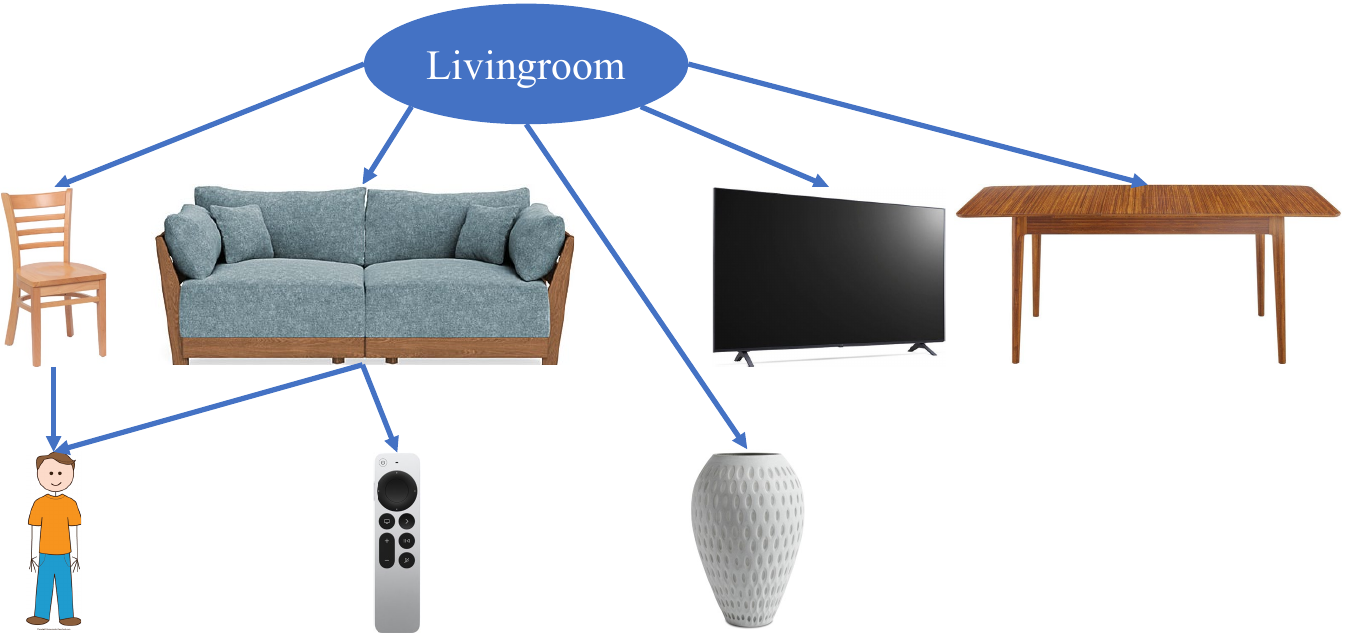}
         \caption{Living room}
     \end{subfigure}
     \begin{subfigure}[b]{0.31\linewidth}
         \centering
         \includegraphics[height=2.7cm]{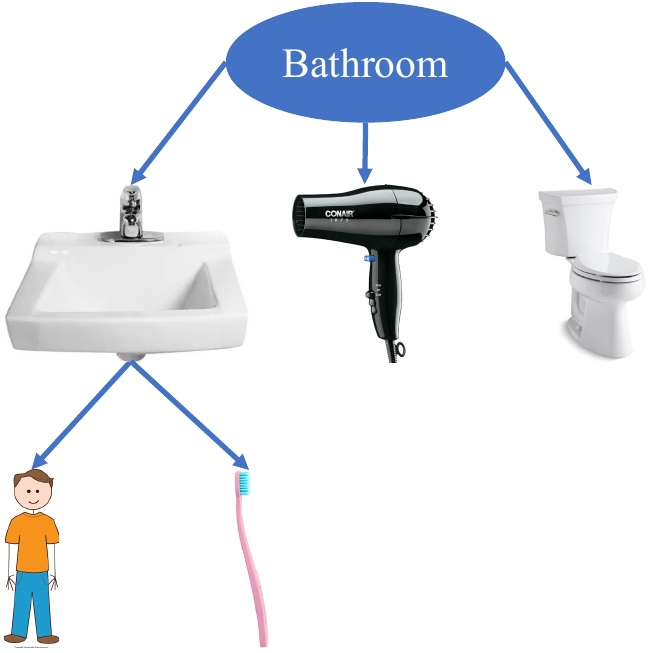}
         \caption{Bathroom}
     \end{subfigure}
      \begin{subfigure}[b]{0.47\linewidth}
         \centering
         \includegraphics[height=2.3cm]{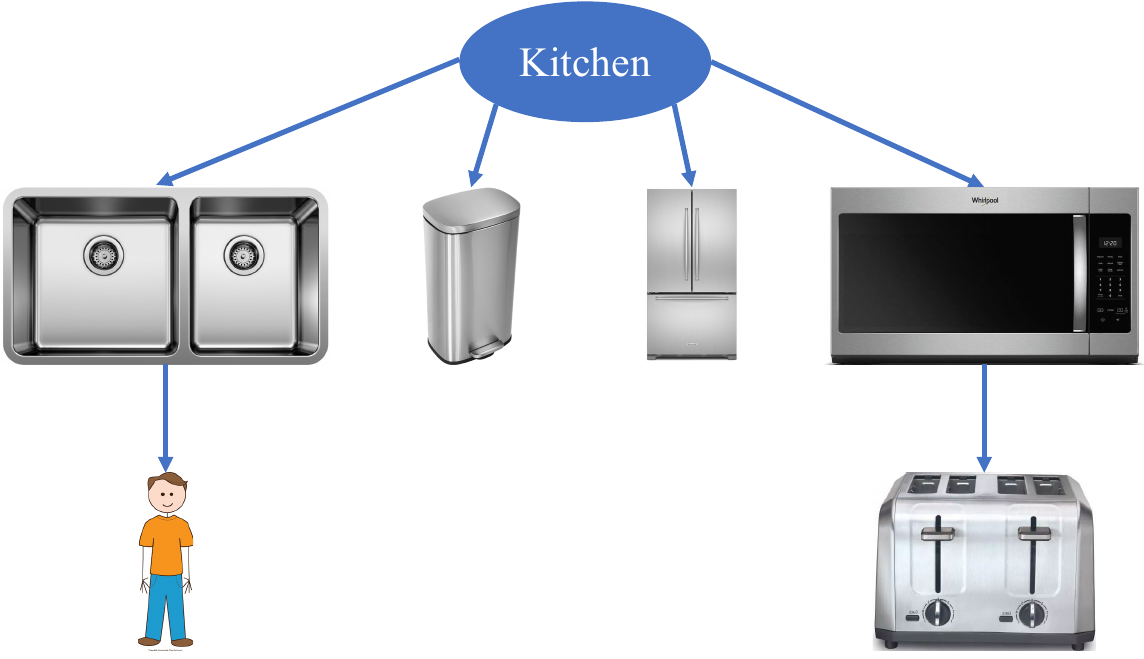}
          \caption{Kitchen}
     \end{subfigure}
     \begin{subfigure}[b]{0.51\linewidth}
         \centering
         \includegraphics[height=2.3cm]{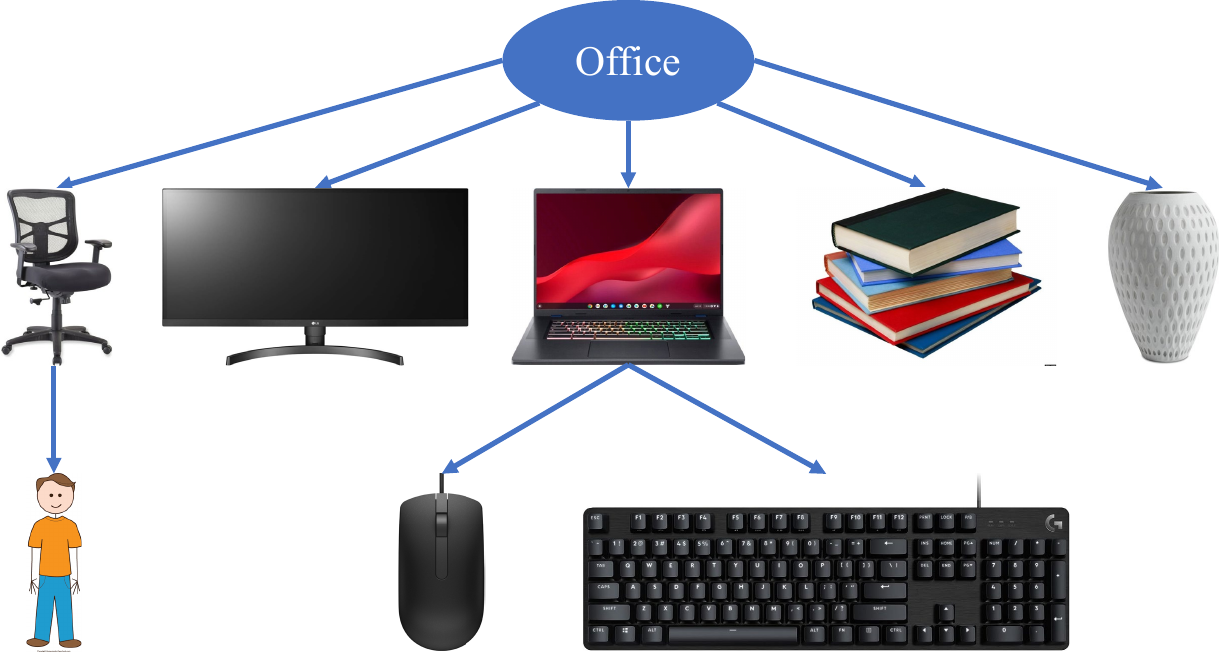}
          \caption{Office}
     \end{subfigure}
    \caption{Bayesian Networks used for four room types in real experiments.}
    \label{fig:bayesian nets}
\end{figure}

 We also conducted a test denying the robot access to semantic prior knowledge. Without semantic prior knowledge,  the robot did not enter the kitchen and failed to find the target, as visualized in the green line in Fig. \ref{fig: kitchen setup}. 

The real-world tests show that our approach runs effectively on a physical robot in unknown real environments, among which the lab environment has many obstacles. It enabled the robot to find the target object nested among obstacles successfully.

\begin{figure}[htbp]
    \centering
\includegraphics[width=0.79\linewidth]{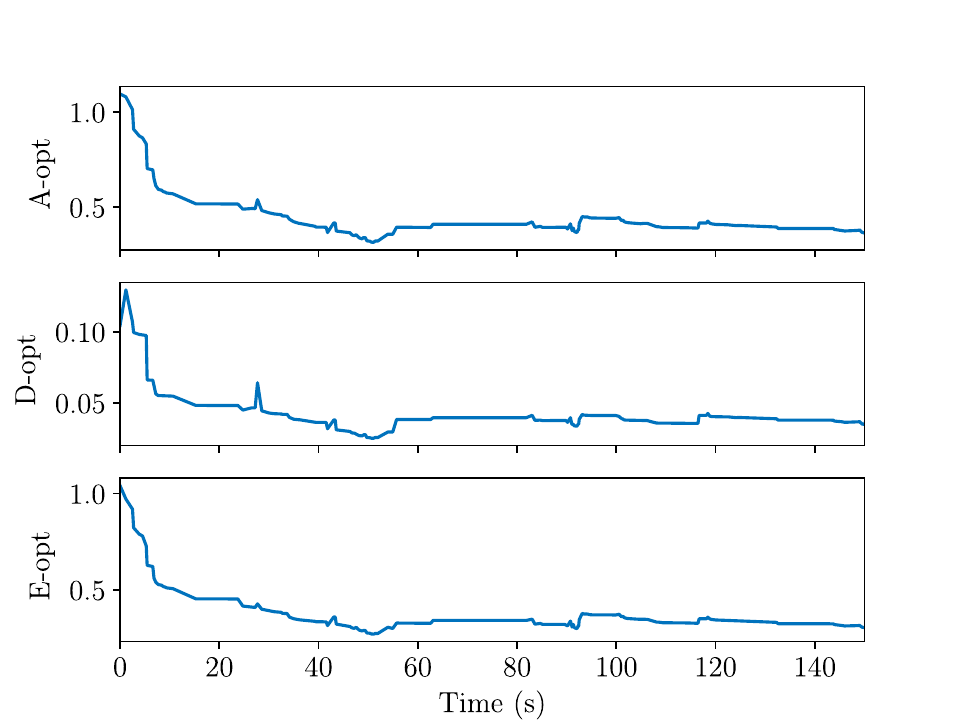}
    \caption{The evolution of the object position covariance over time for all the objects detected in the laboratory environment.}
    \label{fig: real obj covariance}
\end{figure}

We provide the semantic room labels inferred by the Bayesian network in Fig. \ref{fig: semantic room labels}. We can see that at the beginning, due to the lack of evidence, all four room types (``Kitchen'', ``Bathroom'', ``living room'', ``office'') were  equally likely. At the end of the task, the robot correctly concluded that in the laboratory environment, the scene most resembled an office, and in the kitchen environment, the scene was most likely a kitchen. For illustration purposes, we also show the Bayesian networks used for all four room types in Fig. \ref{fig:bayesian nets} without displaying the associated conditional probabilities for brevity.

\begin{figure}
     \centering
     \begin{subfigure}[b]{1\linewidth}
         \centering
         \includegraphics[width=1\textwidth]{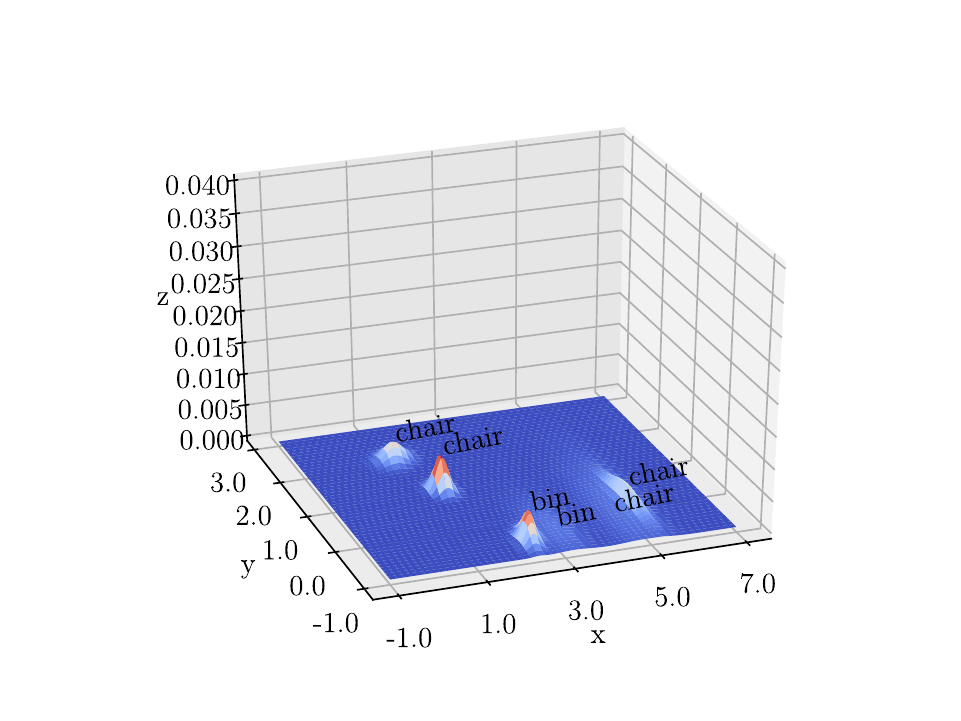}
         \caption{$t=48s$.}
     \end{subfigure}
     \begin{subfigure}[b]{1\linewidth}
         \centering
         \includegraphics[width=1\textwidth]{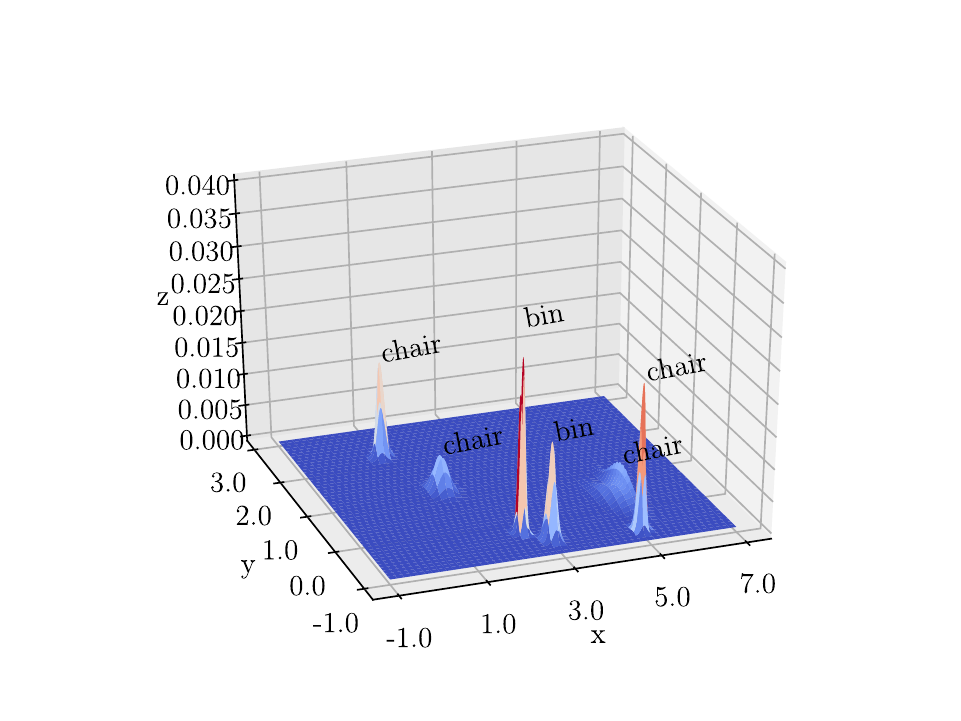}
         \caption{$t=254s$.}
     \end{subfigure}
        \caption{Object position covariance at two time instants in the laboratory environment.}
        \label{fig: real covariance map}
\end{figure}

We now present the semantic SLAM results in the lab demo. Since we do not have the object ground truth information, for the semantic SLAM results, we focus on its uncertainty reduction capability. For all the detected map objects in the lab environment, the A-opt, D-opt, and E-opt of position covariance is calculated. Their evolution over time 
is plotted in Fig. \ref{fig: real obj covariance}. We can see that the position uncertainty of objects decreases with time and is kept at a low level in the end. 

Fig. \ref{fig: real covariance map} gives a more intuitive presentation. Comparing the results at time instants t = 48s
and t = 70s, we can see that at t = 254s, the peak of the bell (which shows the object position covariance) increases, and the bell's base decreases, indicating a more certain
estimation of the object’s position. 

\begin{figure}
    \centering
    \includegraphics[width=0.79\linewidth]{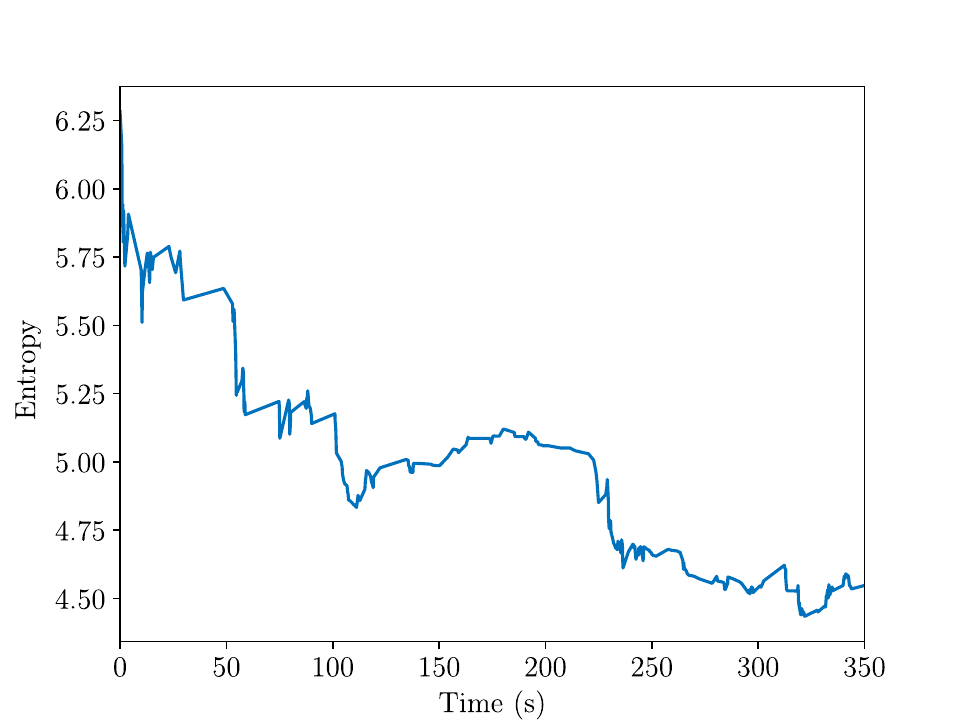}
    \caption{The evolution of the predicted object class entropy with
respect to time for all the objects detected in the laboratory environment.}
    \label{fig: real entropy}
\end{figure}

The entropy of the predicted object class is visualized in Fig. \ref{fig: real entropy}. The result suggests that as time progresses, the robot is more and more certain about the object class it predicted. 

Please refer to the supplementary video for several recorded runs in the virtual and real-world environments. 

% \subsection{Performance Consistency}
% Standard deviations are calculated for the four metrics introduced above: success rate, path length, SPL, and planning time. Since success rate and SPL are calculated by average episodes starting from the same position, the standard deviations of them are calculated by considering the two metrics at different initial positions. On the other hand, the standard deviations of the path length and planning time are computed by running multiple episodes from the same initial position. The calculated standard deviations are summarized in Table \ref{tab:std}. The results suggest that our method has the smallest standard deviations in terms of success rate, path length, and planning time, i.e., it has consistent performance. Our approach shows a larger variation in SPL than the FE baseline, but the FE baseline has a smaller SPL in all episodes we tested and the FE method cannot be used alone to search a semantic object. 

% \begin{table}[htbp]
%     \caption{Standard deviations for all metrics}
%     \centering
%     \begin{tabular}{c c c c c}
%     \toprule
%         Method & Success & Path length (m) & SPL & Planning time (s)\\
%     \midrule 
%     Ours     & \textbf{0.089} & \textbf{1.837} & 0.334 & \textbf{25.01} \\
%     FE & 0.245 & 2.006 & \textbf{0.268} & 29.37\\
%     Ours-NS & 0.363 & 3.617 & 0.415 & 31.41 \\
%     \bottomrule
%     \end{tabular}
%     \label{tab:std}
% \end{table}

\section{Conclusions}
We presented a novel approach to tackle the open problem of enabling
a robot to search for a semantic object in an unknown
and GPS-denied environment. Our approach combines 
semantic SLAM, Bayesian Networks, Markov Decision Process,
and Real-Time Dynamic Programming. The testing and ablation study results demonstrate both the
effectiveness and efficiency of our approach. Moreover, the fused maps produced with our system can be used for performing future tasks in the same environment, and each time, the maps will be enhanced in terms of accuracy and coverage (as the robot explores more of the environment with semantic SLAM) to enable more efficient and effective task performance. 
%Moreover, while our approach is unique in incorporating semantic object information to search for semantic targets, to evaluate its motion planning performance, we compared it to a non-semantic baseline planning method and conducted an ablation study. The results show that our approach has a higher success rate, shorter path length, and less planning time. 
In the next step, we will extend our approach to more complex, compound semantic tasks involving interaction with objects.
%that require the robot to interact with objects. 

% {\appendix[Semantic SLAM Equations]
% \begin{equation*}
% \mathbf{\Psi}^{-1}= \mathbf{K}_2^T\mathbf{\Sigma} _\delta^{-1}\mathbf{K}_2 + \mathbf{\Sigma} _p^{-1}
% \end{equation*}
% \begin{equation*}
% \mu_{p, t} = \begin{bmatrix} \mu_x, \mu_y, \mu_\theta\end{bmatrix}^T
% \end{equation*}
% \begin{equation*}
% \mu_{t} = \begin{bmatrix} \mu_{m,x}, \mu_{m,y}\end{bmatrix}^T
% \end{equation*}
% \begin{equation*}
% \mathbf{K}_1 = \begin{bmatrix}
% \frac{\mu_{m,x} - \mu_x}{\sqrt{(\mu_{m,x} - \mu_x)^2 + (\mu_{m,y} - \mu_y)^2}} & \frac{\mu_{m,y} - \mu_y}{\sqrt{(\mu_{m,x} - \mu_x)^2 + (\mu_{m,y} - \mu_y)^2}} \\
% \frac{\mu_{y} - \mu_{m, y}}{(\mu_{m,x} - \mu_x)^2 + (\mu_{m,y} - \mu_y)^2} & \frac{\mu_{m, x}- \mu_{x}}{(\mu_{m,x} - \mu_x)^2 + (\mu_{m,y} - \mu_y)^2} 
% \end{bmatrix},
% \end{equation*}

% \begin{equation*}
% \mathbf{K}_2 = \begin{bmatrix}
% \frac{\mu_x - \mu_{m,x} }{\sqrt{(\mu_{m,x} - \mu_x)^2 + (\mu_{m,y} - \mu_y)^2}} & \frac{\mu_{m, y} - \mu_{y}}{(\mu_{m,x} - \mu_x)^2 + (\mu_{m,y} - \mu_y)^2} \\
% \frac{\mu_y- \mu_{m,y} }{\sqrt{(\mu_{m,x} - \mu_x)^2 + (\mu_{m,y} - \mu_y)^2}} & \frac{\mu_{x}-\mu_{m, x}}{(\mu_{m,x} - \mu_x)^2 + (\mu_{m,y} - \mu_y)^2} \\
% 0 & -1 \\
% \end{bmatrix}^T.
% \end{equation*}}

% {\appendices
% \section*{Semantic SLAM Equation}
% \section*{Proof of the Second Zonklar Equation}
% Appendix two text goes here.}

\bibliographystyle{IEEEtran}
\bibliography{bibtex/refs}

% \newpage

 

% \end{IEEEbiography}



% \vfill

\end{document}